\pdfoutput = 1
\documentclass{article} % For LaTeX2e
\usepackage{iclr2022_conference,times}

\usepackage{amsmath,amsfonts,bm}

\def\eqref#1{equation~\ref{#1}}
\def\1{\bm{1}}

\DeclareMathAlphabet{\mathsfit}{\encodingdefault}{\sfdefault}{m}{sl}
\SetMathAlphabet{\mathsfit}{bold}{\encodingdefault}{\sfdefault}{bx}{n}

\usepackage{hyperref}
\usepackage{url}

\usepackage[utf8]{inputenc} % allow utf-8 input
\usepackage[T1]{fontenc}    % use 8-bit T1 fonts
\usepackage{url}            % simple URL typesetting
\usepackage{booktabs}       % professional-quality tables
\usepackage{amsfonts}       % blackboard math symbols
\usepackage{nicefrac}       % compact symbols for 1/2, etc.
\usepackage{microtype}      % microtypography
\usepackage{xcolor}         % colors
\usepackage{graphicx,subfigure}
\usepackage{amsmath}
\usepackage{multirow}
\usepackage{longtable}
\usepackage{makecell}
\usepackage{algorithm}
\usepackage{algpseudocode}
\usepackage{listings}

\usepackage{capt-of}
\usepackage{tabularx}

\newcommand{\agent}{\textsc{CoBERL}}
\newcommand{\relic}{\textsc{ReLIC}}

\title{CoBERL: Contrastive BERT for Reinforcement Learning}

\author{%
  Andrea Banino \textsuperscript{1, +}\\
  \And
  Adri\a`a Puidomenech Badia \textsuperscript{1, +}\\
  \And
  Jacob Walker \textsuperscript{1, +}\\
  \AND
  Tim Scholtes \textsuperscript{1, +}\\
  \And
  Jovana Mitrovic \textsuperscript{1}\\
  \And
  Charles Blundell \textsuperscript{1}\\
}

\iclrfinalcopy % Uncomment for camera-ready version, but NOT for submission.
\begin{document}

\let\footnote\relax\footnotetext{\textsuperscript{1} DeepMind London,
\textsuperscript{+} Equal Contribution. 
Corresponding author: \texttt{abanino@deepmind.com}}

\maketitle

\begin{abstract}
Many reinforcement learning (RL) agents require a large amount of experience to solve tasks. 
We propose Contrastive BERT for RL (\agent{}), an agent that combines a new contrastive loss and a hybrid LSTM-transformer architecture to tackle the challenge of improving data efficiency. \agent{} enables efficient and robust learning from pixels across a wide variety of domains. 
We use bidirectional masked prediction in combination with a generalization of a recent contrastive method to learn better representations for RL, without the need of hand engineered data augmentations.
We find that \agent{} consistently improves data efficiency across the full Atari suite, a set of control tasks and a challenging 3D environment, and often it also increases final score performance. 
\end{abstract}

\section{Introduction}
\label{sec:introduction}
Developing sample efficient reinforcement learning (RL) agents that only rely on raw high dimensional inputs is challenging. Specifically, it is difficult as it often requires us to simultaneously train several neural networks based on sparse  environment feedback and strong correlation between consecutive observations. This problem is particularly severe when the networks are large and densely connected, like in the case of the transformer \citep{vaswani2017attention} due to noisy gradients often found in RL problems. 
Outside of the RL domain, transformers have proven to be very expressive \citep{brown2020language} and, as such, they are of particular interest for complex domains like RL. 

In this paper, we propose to tackle this shortcoming by taking inspiration from Bidirectional Encoder Representations for Transformers \citep[BERT; ][]{devlinetal2019bert}, and its successes on difficult sequence prediction and reasoning tasks. Specifically we propose a novel agent, named Contrastive BERT for RL (\agent{}), that combines a new contrastive representation learning objective with architectural improvements that effectively combine LSTMs with transformers.

For representation learning we take inspiration from previous work showing that contrastive objectives improve the performance of agents \citep{fortunato2019generalization, srinivas2020curl, kostrikov2020image, mitrovic2020representation}. 
Specifically, we combine the paradigm of masked prediction from BERT \citep{devlinetal2019bert} with the contrastive approach of Representation Learning via Invariant Causal Mechanisms \citep[\textsc{ReLIC};][]{mitrovic2020representation}.
Extending the BERT masked prediction to RL is not trivial; unlike in language tasks, there are no discrete targets in RL. To circumvent this issue, we extend \relic{} to the time domain and use it as a proxy supervision signal for the masked prediction. Such a signal aims to learn self-attention-consistent representations that contain the appropriate information for the agent to effectively incorporate previously observed knowledge in the transformer weights. 
Critically, this objective can be applied to different RL domains as it does not require any data augmentations and thus circumvents the need for much domain knowledge. This is another advantage compared to the original \relic{} and which require to hand-design augmentations.

In terms of architecture, we base \agent{} on Gated Transformer-XL \citep[GTrXL; ][]{parisotto2020stabilizing} and Long-Short Term Memories \citep[LSTMs; ][]{hochreiter1997long}. GTrXL is an adaptation of a transformer architecture specific for RL domains. We combine GTrXL with LSTMs using a gate trained via an RL loss. This allows the agent to learn to exploit the representations offered by the transformer only when an environment requires it, and avoid the extra complexity when not needed.

We extensively test our proposed agent across a widely varied set of environments and tasks ranging from 2D platform games to 3D first-person and third-person view tasks.
Specifically, we test it in the control domain using DeepMind Control Suite \citep{tassa2018deepmind} and probe its memory abilities using DMLab-30 \citep{beattie2016deepmind}. 
We also test our agent on all 57 Atari games \citep{bellemare2013arcade}. Our main contributions are:

\begin{itemize}
    \item A novel contrastive representation learning objective that combines the masked prediction from BERT with a generalization of \relic{} to the time domain; with this we learn self-attention consistent representations and extend BERT-like training to RL using contrastive objectives without the need of hand-engineered augmentations. 
    \item An improved architecture that, using a gate, allows us to flexibly integrate knowledge from both the transformer and the LSTM.
    \item Improved data efficiency (and in some cases also overall performance) on a varied set of environments. Also, we show that individually both our contrastive loss and the architecture improvements play a role in improving performance.
\end{itemize}
\vspace{-0.5cm}

\begin{figure}[t!]
\label{fig:arch}
\centering
  \includegraphics[scale=0.36]{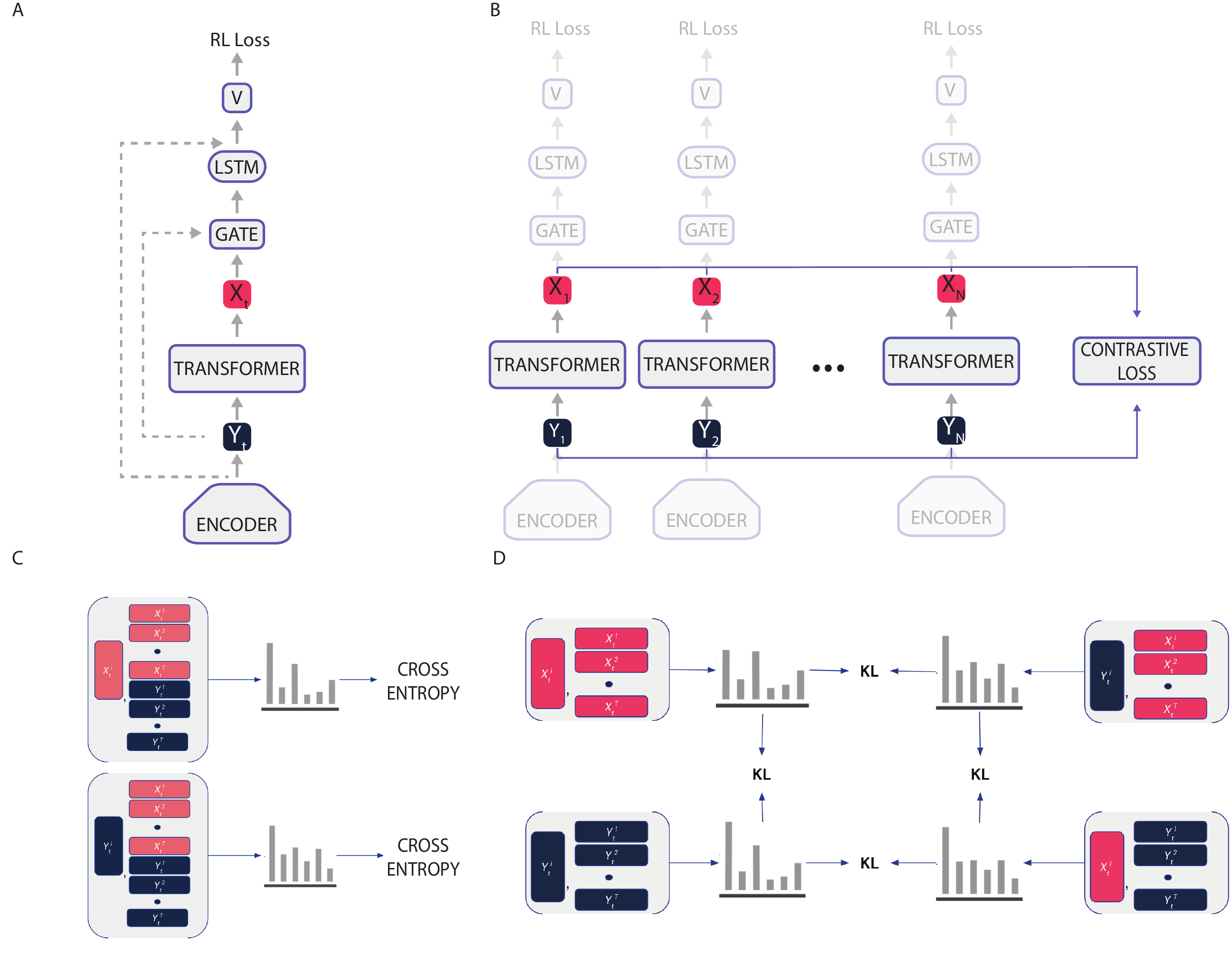}
  \vspace*{-0.5cm}
  \caption{\small{\agent{}. 
  A) General Architecture. We use a residual network to encode observations into embeddings $Y_{t}$.
  We feed $Y_{t}$ through a causally masked GTrXL transformer, which computes the predicted masked inputs $X_{t}$ and passes those together with $Y_{t}$ to a learnt gate. The output of the gate is passed through a single LSTM layer to produce the values that we use for computing the RL loss. B) Contrastive loss. We also compute a contrastive loss using predicted masked inputs $X_{t}$ and $Y_{t}$ as targets. For this, we do not use the causal mask of the Transfomer. For details about the contrastive loss, please see Section \ref{sec:unsupervised-learning}. C) Computation of $q_{x}^{t}$ and $q_{y}^{t}$ (from Equation \ref{eq:main}) with the round brackets denoting the computation of the similarity between the entries D) Regularization terms from Eq.~\ref{eq:loss} which explicitly enforce self-attention consistency.}}
\end{figure}

\section{Method}
\label{sec:method}

To tackle the problem of data efficiency in deep reinforcement learning, we propose two modifications to the status quo. First, we introduce a novel representation learning objective aimed at learning better representations by enforcing self-attention consistency in the prediction of masked inputs. Second, we propose an architectural improvement to combine the strength of LSTMs and transformers. 

\subsection{Representation Learning in reinforcement learning}
\label{sec:unsupervised-learning}

It has been empirically observed that performing RL directly from high dimensional observations (e.g. raw pixels) is sample inefficient \citep{lake2017building}; however, there is evidence that learning from low-dimensional state based features is significantly more efficient \citep[e.g.][]{tassa2018deepmind}. Thus, if state information could be effectively extracted from raw observations it may then be possible to learn from these as fast as from states. However, unlike in the supervised or unsupervised setting, learning representations in RL is complicated by non-stationarity in the training distribution and strong correlation between observations at adjacent timesteps. Furthermore, given the often sparse reward signal coming from the environment, learning representations in RL has to be achieved with little to no supervision. Approaches to address these issues can be broadly classified into two classes. The first class uses auxiliary self-supervised losses to accelerate the learning speed in model-free RL algorithms \citep[][]{schmidhuber1990making, jaderberg2016reinforcement, oord2018representation,  Srinivas2020CURLCU}. The second class learns a world model and uses this to collect imagined rollouts, which then act as extra data to train the RL algorithm reducing the samples required from the environment \citep{sutton1990integrated, ha2018world, kaiser2019model, schrittwieser2020mastering}. \agent{} is part of the first set of methods, as it uses a self-supervised loss to improve data efficiency. 
In particular, we take inspiration from the striking progresses made in recent years in both masked language modelling \citep[BERT,][]{devlinetal2019bert} and contrastive learning \citep{oord2018representation, chen20icml}. 
% Our hypothesis is based on the idea that a temporally extended contrastive learning objective will learn representations that once fed into the RL algorithms will allow for a significant gain in terms of data efficiency. Also, we want to achieve this without using hand-designed data augmentations techniques. To do so we take inspiration from the striking progresses made in recent years in both masked language modelling \citep[BERT,][]{devlinetal2019bert} and contrastive learning \citep{oord2018representation, chen20icml}.

From BERT we borrow the combination of bidirectional processing in transformers (rather than left-to-right or right-to-left, as is common with RNN-based models such as LSTMs) with a masked prediction setup.
With this combination the model is forced to focus on the context provided by the surrounding timesteps to solve its objective \citep{voita2019bottom}. Thus, when the model is asked to reconstruct a particular masked frame it does so by attending to all the relevant frames in the trajectory. This is of particular relevance in RL, where state aliasing is a source of uncertainty in value estimation, and so we believe that attending to other frames in the sequence could be a way to provide extra evidence to reduce this uncertainty. 

However, unlike in BERT where the input is a discrete vocabulary for language learning and targets are available, in RL inputs consist of images, rewards and actions that do not form a finite or discrete set and targets are not available. Thus, we must construct proxy targets and the corresponding proxy tasks to solve. For this we use contrastive learning, and we derive our contrastive loss from \relic{}~\citep{mitrovic2020representation}. \relic{} creates a series of augmentations from the original data and then enforces invariant prediction of proxy targets across augmentations through an invariance regularizer, yielding improved generalization guarantees. Compared to \relic{}, \agent{} does not use data augmentations. Instead we rely on the sequential nature of our input data to create the necessary groupings of similar and dissimilar points needed for contrastive learning. Not having a need for augmentations is critical for RL as each domain would require handcrafting different augmentations. Augmentations also make other methods less data efficient. Finally, we do not use an additional encoder network as in \relic{}, thus making \agent{} fully end-to-end. 

We now set out to explain the details of the auxiliary loss. In a batch of sampled sequences, before feeding embeddings into the transformer stack, $15\%$ of the embeddings are replaced with a fixed token denoting masking. Then, let the set $\mathcal{T}$ represent indices in the sequence that have been randomly masked and let $t\in\mathcal{T}$.
For the $i$-th training sequence in the batch, for each index $t\in\mathcal{T}$, let $x_{t}^{i}$ be the output of the GTrXL and $y_{t}^{i}$ the corresponding input to the GTrXL from the encoder (see Fig~\ref{fig:arch}B).
Let $\phi(\cdot, \cdot)$ be the inner product defined on the space of critic embeddings, i.e. $\phi(x,y) = g(x)^T g(y)$, where $g$ is a critic function. The critic separates the embeddings used for the contrastive proxy task and the downstream RL task. Details of the critic function are in App.~\ref{app:architecture}.
This separation is needed since the proxy and downstream tasks are related but not identical, and as such the appropriate representations will likely not be the same.
As a side benefit, the critic can be used to reduce the dimensionality of the dot-product.
To learn the embedding $x_{t}^{i}$ at mask locations $t$, we use $y_{t}^{i}$ as a positive example and the sets $\{y^{b}_{t}\}_{b=0, b\neq i}^{B}$ and $\{x^{b}_{t}\}_{b=0,  b\neq i}^{B}$ as the negative examples with $B$ the number of sequences in the minibatch. 
We model $q^{t}_{x}$ as
\begin{equation}\label{eq:main}
q^{t}_{x} = \frac{\exp(\phi(x_{t}^{i}, y_{t}^{i}))}{\sum_{b=0}^{B}\exp(\phi(x_{t}^{i}, y^{b}_{t})) + \exp(\phi(x_{t}^{i}, x^{b}_{t}))}
\end{equation}
with $q^{t}_{x}$ denoting $q^{t}_{x}\left(x_{t}^{i}\vert \{y^{b}_{t}\}_{b=0}^{B}, \{x^{b}_{t}\}_{b=0,  b\neq i}^{B}\right)$; $q^{t}_{y}$ is computed analogously (see Fig~\ref{fig:arch}C).
In order to enforce self-attention consistency in the learned representations, we explicitly regularize the similarity between the pairs of transformer embeddings and inputs through Kullback-Leibler regularization from \relic{}. 
Specifically, we look at the similarity between appropriate embeddings and inputs, and within the sets of embeddings and inputs separately.
To this end, we define:

\begin{equation} \label{equa:1}
p^{t}_{x} = \frac{\exp(\phi(x_{t}^{i}, y_{t}^{i}))}{\sum_{b=0}^{B}\exp(\phi(x_{t}^{i}, y^{b}_{t}))};
\quad
s^{t}_{x} = \frac{\exp(\phi(x_{t}^{i}, x_{t}^{i}))}{\sum_{b=0}^{B}\exp(\phi(x_{t}^{i}, x^{b}_{t}))}
\end{equation} 

with $p^{t}_{x}$ and $s^{t}_{x}$ shorthand for $p^{t}_{x}(x_{t}^{i}\vert \{y^{b}_{t}\}_{b=0}^{B})$ and $s^{t}_{x}(x_{t}^{i}\vert \{x^{b}_{t}\}_{b=0}^{B})$, respectively; $p^{t}_{y}(y_{t}^{i}|\{x^{b}_{t}\}_{b=0}^{B})$ and $s^{t}_{y}(y_{t}^{i}|\{y^{b}_{t}\}_{b=0}^{B})$ defined analogously (see Fig~\ref{fig:arch}D).
All together, the final objective takes the form:

\begin{multline}
\label{eq:loss}
    \mathcal{L}(X, Y) = -\sum_{t\in \mathcal{T}}{} (\log q^{t}_{x} + \log q^{t}_{y})  + \alpha \sum_{t\in \mathcal{T}}{} 
\Big[
KL(s^{t}_{x}, \textrm{sg}(s^{t}_{y})) + KL(p^{t}_{x}, \textrm{sg}(p^{t}_{y}))  +\\ KL(p^{t}_{x}, \textrm{sg}(s^{t}_{y})) +  KL(p^{t}_{y}, \textrm{sg}(s^{t}_{x})) \Big]
\end{multline}
with $\textrm{sg}(\cdot)$ indicating a stop-gradient. As in our RL objective, we use the full batch of sequences that are sampled from the replay buffer to optimize this contrastive objective. Finally, we optimize a weighted sum of the RL objective and $\mathcal{L}(X, Y)$ (see App.~\ref{app:hyperparameters} for the details on the weighting).

\subsection{Architecture of \agent{}.}
\label{sec:arch-improv}

While transformers have proven very effective at connecting long-range data dependencies in natural language processing~\citep{vaswani2017attention, brown2020language, devlinetal2019bert} and computer vision~\citep{carion2020endtoend, dosovitskiy2021an}, in the RL setting they are difficult to train and are prone to overfitting ~\citep{parisotto2020stabilizing}. In contrast, LSTMs have long been demonstrated to be useful in RL. Although less able to capture long range dependencies due to their sequential nature, LSTMs capture recent dependencies effectively. We propose a simple but powerful architectural change: we add an LSTM layer on top of the GTrXL with an extra gated residual connection between the LSTM and GTrXL, modulated by the input to the GTrXL (see Fig~\ref{fig:arch}A). Finally we also have a skip connection from the transformer input to the LSTM output.

More concretely, let $Y_t$ be the output of the encoder network at time $t$, then the additional module can be defined by the following equations (see Fig~\ref{fig:arch}A, $\textrm{Gate}()$, below, has the same form as other gates internal to GTrXL):

\begin{equation}
    X_t = \textrm{GTrXL}(Y_t); \quad Z_t = \textrm{Gate}(Y_t, X_t); \quad \textrm{Output}_t = \textrm{concatenate}(\textrm{LSTM}(Z_t), Y_t)
    \label{eq:arch}
\end{equation}

% These modules are complementary as the transformer has no recency bias \citep{ravfogel2019studying}, whilst the LSTM is biased to represent more recent inputs - the gate in equation \ref{eq:arch} allows this to be a mix of encoder representations and transformer outputs. 

% This memory architecture is agnostic to the choice of RL regime and we evaluate this architecture in both the on and off-policy settings. For on-policy, we use V-MPO\citep{song2019v} as our RL algorithm. V-MPO uses a target distribution for policy updates, and partially moves the parameters towards this target subject to KL constraints. For the off-policy setting, we use R2D2 \citep{kapturowski2018recurrent} which adapts replay and the RL learning objective for agents with recurrent architectures, such as LSTMs, GTrXL, and \agent{}. (please see further details in Appendix \ref{app:extra-discussion} and the pseudo-code in Appenix \ref{app:pseudo-code})

The architecture of \agent{} is based on the idea that LSTMs and Transformer can be complementary. In particular, the LSTM is known for having a short contextual memory. However, by putting a transformer before the LSTM, the embeddings provided to the LSTM have already benefited from the ability of the Transformer  to process long contextual dependencies, thus helping the LSTM in this respect. The second benefit works in a complementary direction. Transformers suffer from a quadratic computational complexity with respect to the sequence length. Our idea is that by letting an LSTM process some of this contextual information we can reduce the memory size of the Transformer up to the point where we see a loss in performance. This hypothesis come from studies showing that adding recurrent networks in architectures has the ability to closely emulate the behavior of non-recurrent but deeper models, but it does so with far fewer parameters \citep{schwarzschild2021uncanny}. Having fewer parameters is an aspect of particular interest in RL, where gradients are noisy and hence training larger model is more complicated than supervised learning \citep{mccandlish2018empirical}. Both hypotheses have been empirically confirmed by our ablations in section \ref{sec:ablations}. 

The learnt gate (eq. \ref{eq:arch} and Appendix \ref{app:gate}), was done to give \agent{} the ability to initially skip the output of the transformer stack until the weights of this are warmed-up. Once the transformer starts to output useful information the agent will be able to tune the weights in the gate to learn from them. We hypothesize that this would also help in terms of data efficiency as at the beginning of training the agent could only use the LSTM to get off the ground and start collecting relevant data to train the Transformer. Finally, the skip connection on the output of the LSTM was taken from \cite{kapturowski2018recurrent}.

% For skip connections, the one between the GTrXL module and the LSTM, was done to give the ability to \agent{} to initially skip the output of the Transformer stack until the weights of this were warmed-up. We implemented this with learnt gate (eq. \ref{eq:arch}) with the hypothesis that, once the transformer starts to output useful information then the agent will be able to tune the weights in the gate to learn from them. We figured this would also help in terms of data efficiency as at the beginning of training the agent could quickly use the LSTM to get off the ground to start collecting interesting data. 
% \agent{} is indeed better in data efficiency as shown by Table \ref{tab:atari_results} (5th and 25th percentile), Figure \ref{fig:dm_lab_plots}B, and Appendix \ref{app:learning-curves}.

Since the architecture is agnostic to the choice of RL regime we evaluate it in both on-policy and off-policy settings. For on-policy, we use V-MPO \citep{song2019v}, and for off-policy we use R2D2 \citep{kapturowski2018recurrent}.

\paragraph{R2D2 Agent}
Recurrent Replay Distributed DQN~\citep[R2D2;][]{kapturowski2018recurrent} demonstrates how replay and the RL learning objective can be adapted to work well for agents with recurrent architectures. Given its competitive performance on Atari-57, we implement our CoBERL architecture in the context of Recurrent Replay Distributed DQN~\citep{kapturowski2018recurrent}. We effectively replace the LSTM with our gated transformer and LSTM combination and add the contrastive representation learning loss. With R2D2 we thus leverage the benefits of distributed experience collection, storing the recurrent agent state in the replay buffer, and "burning in" a portion of the unrolled network with replayed sequences during training. 
\vspace{-0.4cm}
\paragraph{V-MPO Agent}
Given V-MPO's strong performance on DMLab-30, in particular in conjunction with the GTrXL architecture \citep{parisotto2020stabilizing} which is a key component of CoBERL, we use V-MPO and DMLab-30 to demonstrate CoBERL's use with on-policy algorithms.
V-MPO is an on-policy adaptation of Maximum a Posteriori Policy Optimization (MPO)~\citep{abdolmaleki2018relative}. To avoid high variance often found in policy gradient methods, V-MPO uses a target distribution for policy updates, subject to a sample-based KL constraint, and gradients are calculated to partially move the parameters towards the target, again subject to a KL constraint. Unlike MPO, V-MPO uses a learned state-value function $V(s)$ instead of a state-action value function.

\section{Related Work}
\label{sec:related_work}

% \paragraph{Long-term memory in RL.}
% The use of memory in deep reinforcement learning has recently been a frequent subject of study. 
% Methods in this area usually combine architectural modifications to existing agents as to enforce learning long-range dependencies. 
% Agents such as MERLIN~\citep{wayne2018unsupervised} and MRA~\citep{fortunato2019generalization} combine existing RL policy and value networks with an augmented external memory and LSTMs. 
% MRA uses a slot-based external memory, akin to memory networks \citep{weston2014memory} and similar models of episodic memory \citep{pritzel2017neural}, and incorporates one of the early contrastive methods CPC~\citep{oord2018representation} as an auxiliary loss.
% On the other hand, MERLIN uses a fully parametric model, the Differential Neural Computer~\citep{graves2016hybrid}, and a variational autoencoder objective to shape embeddings.  
% Although these approaches have made significant progress in advancing long-term memory in RL, they do not effectively incorporate memories across multiple scales and are hugely data-inefficient requiring up to $100$ billion frames of experience to make progress on tasks requiring reasoning over multiple timescales.

% not sure why this is relevant: gupta2017cognitive seems better for when we have the city experiments.
% dreamer is a strong baseline on these domains, but is very different otherwise.
% \textcolor{red}{mapping an environment~\citep{gupta2017cognitive}ch
% We should include the dreamer paper in one of these, since we use it as a baseline in experiments}

\paragraph{Transformers in RL.}
The transformer architecture \citep{vaswani2017attention} has recently emerged as one of the best performing approaches in language modelling \citep{dai2019transformer, brown2020language} and question answering \citep{dehghani2018universal, yang2019xlnet}. 
More recently it has also been successfully applied to computer vision \citep{dosovitskiy2021an}.
Given the similarities of sequential data processing in language modelling and reinforcement learning, transformers have 
also been successfully applied to the RL domain, where as motivation for GTrXL, \cite{parisotto2020stabilizing} noted that extra gating was helpful to train transformers for RL due to the high variance of the gradients in RL relative to that of (un)supervised learning problems. 
In this work, we build upon GTrXL and demonstrate that, perhaps for RL: attention is not all you need, and by combining GTrXL in the right way with an LSTM, superior performance is attained.
We reason that this demonstrates the advantage of both forms of memory representation: the all-to-all attention of transformers combined with the sequential processing of LSTMs.
In doing so, we demonstrate that care should be taken in how LSTMs and transformers are combined and show a simple gating is most effective in our experiments. 
Also, unlike GTrXL, we show that using an unsupervised representation learning loss that enforces self-attention consistency is an effective way to enhance data efficiency when using transformers in RL.
\vspace{-0.4cm}
\paragraph{Contrastive Learning.} 
Recently contrastive learning \citep{hadsell2006dimensionality,  gutmann2010noise, oord2018representation} has emerged as a very performant paradigm for unsupervised representation learning, in some cases even surpassing supervised learning \citep{chen20icml, Caron2020UnsupervisedLO, mitrovic2020representation}. 
These methods have also been leveraged in an RL setting with the hope of improving performance.
Apart from MRA~\citep{fortunato2019generalization} mentioned above, one of the early examples of this is CURL \citep{Srinivas2020CURLCU} which combines Q-Learning with a separate encoder used for representation learning with the InfoNCE loss from CPC \citep{oord2018representation}. 
More recent examples use contrastive learning for predicting future latent states \citep{Schwarzer2020DataEfficientRL, Mazoure2020DeepRA}, defining a policy similarity embeddings \citep{Agarwal2021ContrastiveBS} and learning abstract representations of state-action pairs \citep{liu2021returnbased}. The closest work to our is M-CURL \citep{Zhu2020MaskedCR}. Like our work, it combines mask prediction, transformers, and contrastive learning, but there are a few key differences. First, unlike M-CURL who use a separate policy network, \agent{} computes Q-values based on the output of the transformer. Second, \agent{} combines the transformer architecture with a learnt gate (eq. \ref{eq:gate-def} in App. \ref{app:gate} to produce the input for the Q-network, while M-CURL uses the transformer as an additional embedding network (critic) for the computation of the contrastive loss. Third, while \agent{} uses an extension of \relic{} \citep{mitrovic2020representation} to the time domain and operates on the inputs and outputs of the transformer, M-CURL uses CPC \citep{oord2018representation} with a momentum encoder as in \citep{Srinivas2020CURLCU} and compares encodings from the transformer with the separate momentum encoder.
\vspace{-0.4cm}
\section{Experiments}

% \textcolor{red}{Steven: A general observation I have is that there are no learning curves anywhere in the paper, just tables with final performance. I think this makes it harder to understand what the improvement to sample efficiency actually is, especially when final performance is comparable across different methods}
\vspace{-0.3cm}
\label{sec:experiments}
We provide empirical evidence to show that \agent{} i) improves data efficiency across a wide range of environments and tasks, and ii) needs all its components to maximise its performance. 
In our experiments, we demonstrate performance on Atari57~\citep{bellemare2013arcade}, the DeepMind Control Suite~\citep{tassa2018deepmind}, and the DMLab-30 \citep{beattie2016deepmind}.
Recently, Dreamer V2 \citep{hafner2020mastering} has emerged as a strong model-based agent across Atari57 and DeepMind Control Suite; we therefore include it as a reference point for performance on these domains.

For all experiments, we report scores at the end of training. All results are averaged over five seeds and reported with standard error. For a more comprehensive description of the evaluation done, as well as detailed information on the distributed setup used see App.~\ref{app:distributed_R2D2}. Ablations are run on 7 Atari games chosen to match the ones in the original DQN publication \citep{mnih2013playing}, and on all the 30 DMLab games. The hyper-parameters of all the baselines are tuned individually to maximise performance (see App.~\ref{app:hypers_process} for the detailed procedure). We use a ResNet as the encoder for \agent{}. We use Peng's $Q(\lambda)$ as our loss \citep{peng1994incremental}. To ensure that this is comparable to R2D2, we also run an R2D2 baseline with this loss. For thorough description of the architectures for all environments see App.~\ref{app:architecture}.

For DMLab-30 we use V-MPO \citep{song2019v} to directly compare \agent{} with \citep{parisotto2020stabilizing} and also demonstrate how \agent{} may be applied to both on and off-policy learning. The experiments were run using a Podracer setup \citep{hessel2021podracer}, details of which may be found in App.~\ref{app:distributed_VMPO}. \agent{} is trained for 10 billion steps on all 30 DMLab-30 games at the same time, to mirror the exact multi-task setup presented in \citep{parisotto2020stabilizing}. Compared to \citep{parisotto2020stabilizing} we have two differences. Firstly, all the networks run without pixel control loss \citep{jaderberg2016reinforcement} so as not to confound our contrastive loss with the pixel control loss. Secondly all the models used a fixed set of hyperparameters with 3 random seeds, whereas in \citep{parisotto2020stabilizing} the results were averaged across hyperparameters.

\subsection{Testing \agent{} across several environments}
% Here should be all the results of the final agent.

To test the generality of our approach, we analyze the performance of our model on a wide range of environments. We show results on the Arcade Learning Environment~\citep{bellemare2013arcade}, DeepMind Lab~\citep{beattie2016deepmind}, as well as the DeepMind Control Suite~\citep{tassa2018deepmind}. 
To help with comparisons, in Atari-57 and DeepMind Control we introduce an additional baseline, which we name R2D2-GTrXL. In this variant of R2D2 the LSTM is replaced by GTrXL. R2D2-GTrXL has no unsupervised learning. This way we are able to observe how GTrXL is affected by the change to an off-policy agent (R2D2), from its original V-MPO implementation in~\cite{parisotto2020stabilizing}. We also perform an additional ablation analysis by removing the contrastive loss from \agent{} (see Sec. \ref{sec:aux-loss-ablations}). This baseline demonstrates the importance of contrastive learning in these domains, and we show that the combination of an LSTM and transformer is superior to either alone. 

\vspace{-0.7cm}
\begin{minipage}{0.70\textwidth}
\begin{table}[H]
    \centering
    \footnotesize
    % \begin{tabular}{p{1.2cm}|p{0.3cm}|p{1.1cm}|p{1.0cm}|p{1.0cm}|@{\hspace{0.5mm}}c}
    \begin{tabular}{r@{\hspace{1mm}}|@{\hspace{0.3mm}}c@{\hspace{1mm}}|@{\hspace{0.4mm}}c@{\hspace{1mm}}|@{\hspace{0.5mm}}c@{\hspace{1mm}}|@{\hspace{0.5mm}}c@{\hspace{1mm}}|@{\hspace{0.5mm}}c}
      & $>$ h. & Mean & Median & 25th Pct & 5th Pct \\ \hline
     \agent{} & \textbf{49} & $\mathbf{1368.37\%\pm 34.10\%}$& \textbf{296.22\%} & \textbf{150.56\%} & \textbf{13.65\%}\\ %\hline
     R2D2-GTrXL & 48 &  1088.90\% $\pm$ 53.45\% & 273.83\% & 141.66\% & 5.94\% \\ %\hline
     R2D2 & 47 & 1017.18\% $\pm$ 33.85\% & 269.26\% & 118.96\% & 4.18\% \\ %\hline
     Rainbow & 43 & 874.0\% & 231.0\% & 101.7\% & 4.9\% \\
     Dreamer V2* & 37 & 631.1\% & 162.0\% & 76.6\% & 2.5\%  \\ %\hline
    \end{tabular}
    \caption{\small{
    The human normalized scores on Atari-57. 
    % Normalized scores are computed as $\frac{agent-random \;score}{human-random \;score}$. 
    $>$ h indicates the number of tasks for which performance above average human was achieved.
    $*$ indicates that it was run on 55 games with sticky actions; Pct refers to percentile.}}
    \label{tab:atari_results}
\end{table}
\end{minipage}
\hfill
\begin{minipage}{0.25\textwidth}
\begin{figure}[H]
    \centering
    \includegraphics[scale=0.30]{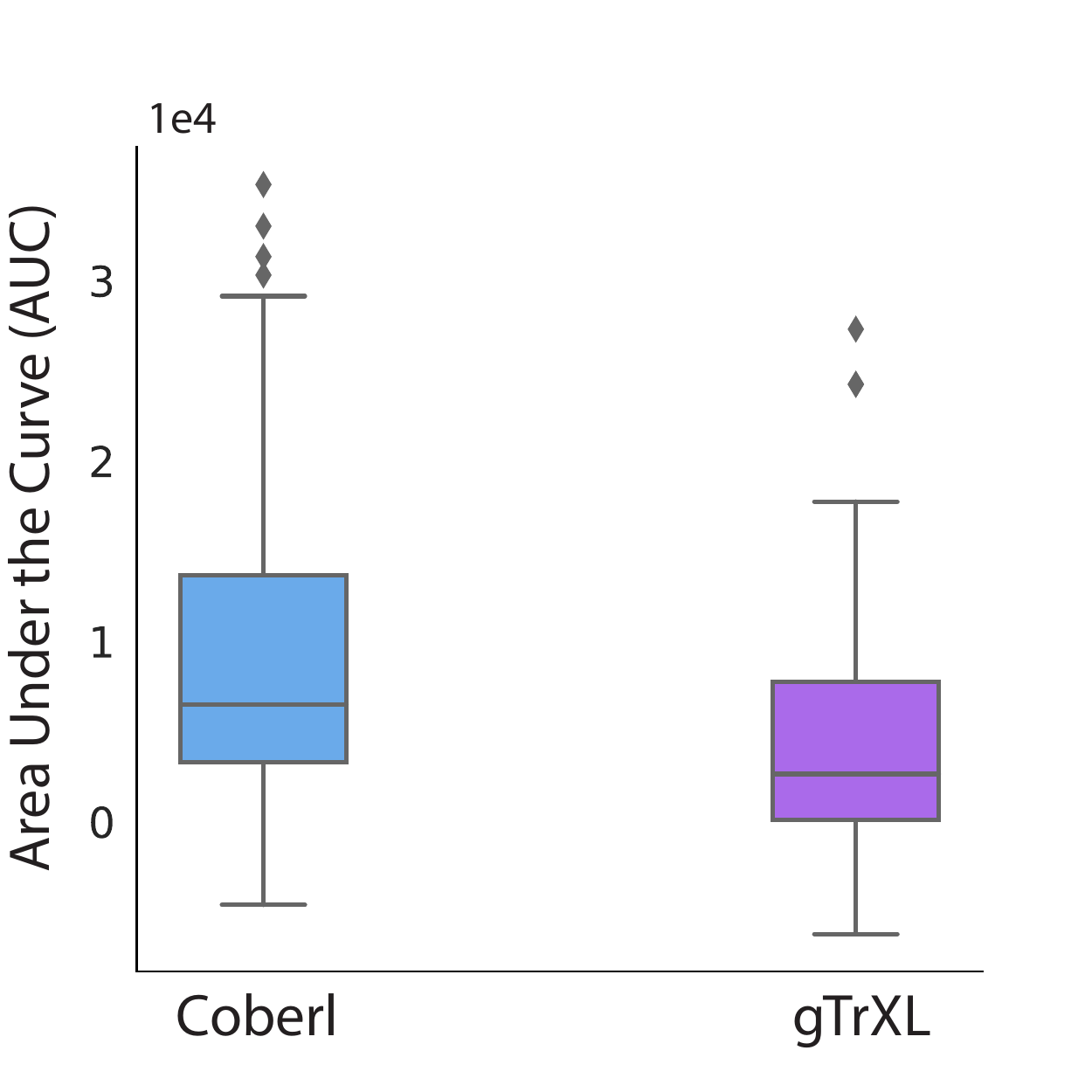}
    \vspace*{-0.8cm}
    \caption{\small{Area under the curve (AUC) for Atari on all 57 levels (higher is better).}} 
    \label{fig:auc-atari}
\end{figure}
\end{minipage}

\paragraph{Atari}
As commonly done in literature~\citep{mnih2015human, hessel2018rainbow, machado2018revisiting, hafner2020mastering}, 
we measure performance on all 57 Atari games after running for $200$ million frames. As detailed in App.~\ref{app:hyperparameters}, we use the standard Atari frame pre-processing to obtain the $84$x$84$ gray-scaled frames that are used as input to our agent. We do not use frame stacking.

Table~\ref{tab:atari_results} shows the results of all the agents where published results are available. \agent{} shows the most games above average human performance and significantly higher overall mean performance. Interestingly, the performance of R2D2-GTrXL shows that the addition of GTrXL is not sufficient to obtain the improvement in performance that \agent{} exhibits (in \ref{sec:ablations} we will demonstrate that both the contrastive loss and LSTM contribute to this improvement). R2D2-GTrXL also exhibits slightly better median than \agent{}, showing that R2D2-GTrXL is indeed a powerful variant on Atari.  Additionally, we observe that the difference in performance in \agent{} is higher when examining the lower percentiles. This suggests that \agent{} causes an improvement in data efficiency. To confirm this tendency we calculated the area under the curve (AUC) for the learning curves presented in Appendix \ref{app:learning-curves}  and we perform a t-test analysis on it (please see App.\ref{app:auc} for details on how AUC was calculated). Figure \ref{fig:auc-atari} present this analysis, and it confirms a significant difference with respect to the model used, t(114)=3.438, p=.008, with \agent{}(M=7380.23, SD=942.29) being better than GTrXL (M=5192.76, SD=942.01), thus showing that \agent{} is more data efficient.

\paragraph{Control Suite}

We also perform experiments on the DeepMind Control Suite~\citep{tassa2018deepmind}. While the action space in this domain is typically treated as continuous, we discretize the action space in our experiments to apply the same architecture as in Atari and DMLab-30. For more details on the number of actions for each task see App.~\ref{app:control_discretization}. We do not use pre-processing on the environment frames. Finally, \agent{} is trained only from pixels without state information. 

\begin{minipage}{0.65\textwidth}
\begin{table}[H]
% \resizebox{\textwidth}{!}{%
\tiny
\begin{tabular}{c|c|c|c|c}
\hline
 DM Suite & \agent{} & R2D2-GTrXL & R2D2&  D4PG-Pixels\\
\hline
 acrobot swingup & \bf{355.25 $\pm$ 3.82} & 265.90 $\pm$ 113.27 & 327.16 $\pm$ 5.35 & 81.7 $\pm$ 4.4 \\
 fish swim & \bf{597.66 $\pm$ 60.09} & 82.30 $\pm$ 265.07 & 345.63 $\pm$ 227.44 & 72.2 $\pm$ 3.0 \\
 fish upright & \bf{952.60 $\pm$ 5.16} & 844.13 $\pm$ 21.70 & 936.09 $\pm$ 11.58 & 405.7 $\pm$ 19.6 \\
 pendulum swingup & \bf{835.06 $\pm$ 9.38} & 775.72 $\pm$ 50.06 & 831.86 $\pm$ 61.54 & 680.9 $\pm$ 41.9 \\
 swimmer swimmer6 & \bf{419.26 $\pm$ 48.37} & 206.95 $\pm$ 58.37 & 329.61 $\pm$ 26.77 & 194.7 $\pm$ 15.9\\
 finger spin & \bf{985.96 $\pm$ 1.69} & 983.74 $\pm$ 10.23 & 980.85 $\pm$ 0.67 & \bf{985.7 $\pm$ 0.6}\\
 reacher easy & \bf{984.00 $\pm$ 2.76} & \bf{982.60 $\pm$ 2.24} & \bf{982.28 $\pm$ 9.30} & 967.4 $\pm$ 4.1\\
 cheetah run & \bf{523.36 $\pm$ 41.38} & 105.23 $\pm$ 144.82 & 365.45 $\pm$ 50.40 & \bf{523.8 $\pm$ 6.8} \\
 walker walk & 781.16 $\pm$ 26.16 & 584.50 $\pm$ 70.28 & 687.18 $\pm$ 18.15 & \bf{968.3 $\pm$ 1.8} \\
 ball in cup catch & 979.10 $\pm$ 6.12 & 975.26 $\pm$ 1.63 & \bf{980.54 $\pm$ 1.94} & 980.5 $\pm$ 0.5 \\
 cartpole swingup & 812.01 $\pm$ 7.61 & 836.01 $\pm$ 4.37 & 816.23 $\pm$ 2.93 & \bf{862.0 $\pm$ 1.1}\\
 cartpole swingup sparse & 714.80 $\pm$ 16.18 & 743.71 $\pm$ 9.27 & \bf{762.57 $\pm$ 6.71}  & 482.0 $\pm$ 56.6\\
\hline
\end{tabular}
\vspace{-0.3cm}
\caption{\small{Results on tasks in the DeepMind Control Suite. CoBERL, R2D2-GTrXL, R2D2, and D4PG-Pixels are trained on 100M frames. In the appendix (\autoref{tab:control_tab_baselines_full}), we show three other approaches as reference and not as a directly comparable baseline.)}}
\label{tab:control_tab_baselines}
% }
\end{table}
\end{minipage}
\hfill
\begin{minipage}{0.30\textwidth}
\begin{figure}[H]
    \vspace{-0.7cm}
    \centering
    \includegraphics[scale=0.35]{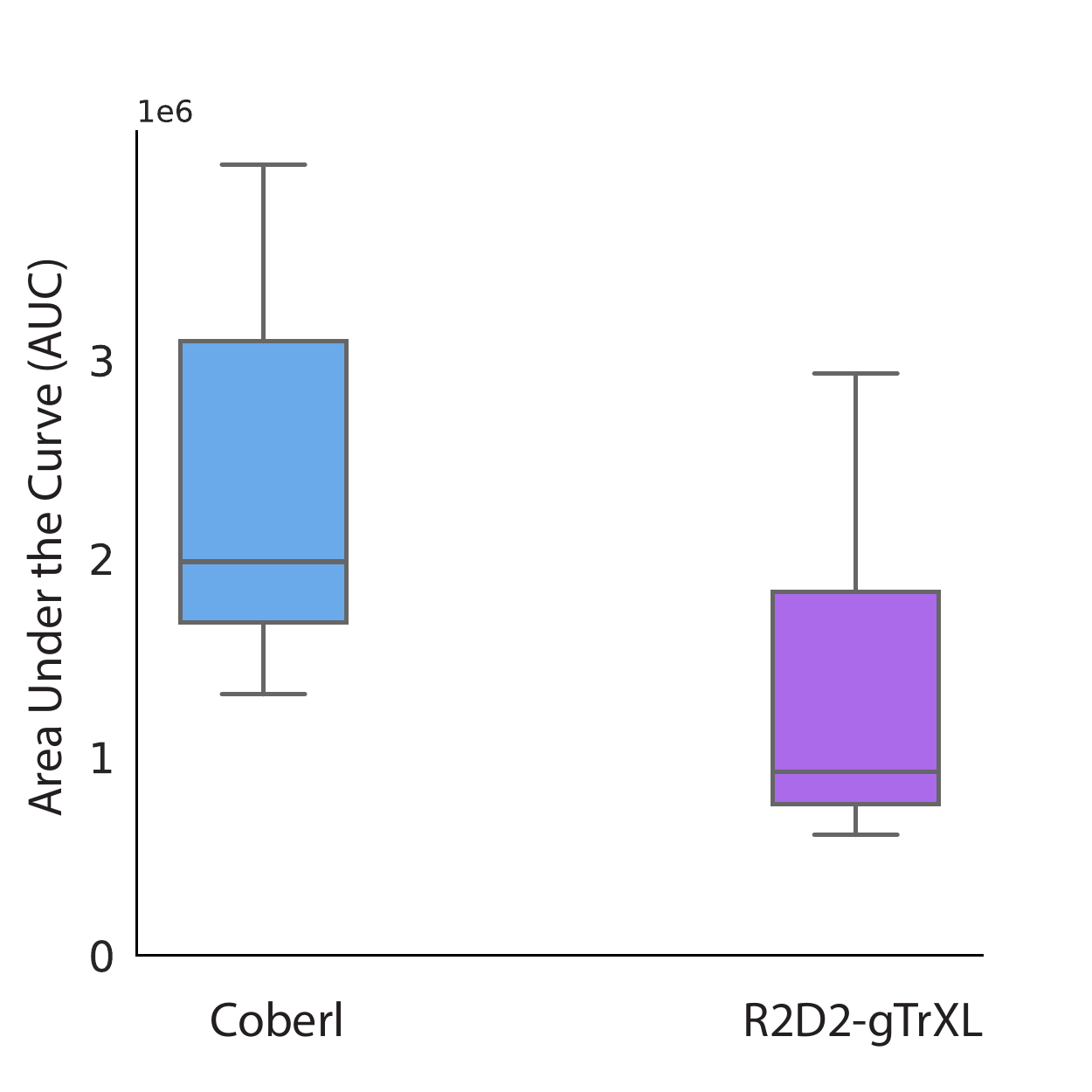}
    \vspace*{-0.8cm}
    \caption{\small{Area under the curve (AUC) for DmControl (higher is better)}.} 
    \label{fig:auc-control}
\end{figure}
\end{minipage}

\vspace{0.1cm}
We include six tasks popular in current literature: {\ttfamily ball\_in\_cup catch}, {\ttfamily cartpole swingup}, {\ttfamily cheetah run}, {\ttfamily finger spin}, {\ttfamily reacher easy}, and {\ttfamily walker walk}.  Most previous work on these specific tasks has emphasized data efficiency as most are trivial to solve even with the baseline---D4PG-Pixels---in the original dataset paper~\citep{tassa2018deepmind}. We thus include 6 other tasks that are difficult to solve with D4PG-Pixels and are relatively less explored: {\ttfamily acrobot swingup}, {\ttfamily cartpole swingup\_sparse}, {\ttfamily fish swim}, {\ttfamily fish upright}, {\ttfamily pendulum swingup}, and {\ttfamily swimmer swimmer6}. 
% We show our results in Figure~\ref{fig:control} as well as a detailed plot on select tasks in Figure~\ref{fig:control_plot}. 
In Table~\ref{tab:control_tab_baselines} we show results on \agent{}, R2D2-gTRXL, R2D2, CURL~\citep{Srinivas2020CURLCU}, Dreamer~\citep{Hafner2020Dream}, Soft Actor Critic~\citep{HaarnojaZAL18} on pixels as demonstrated in~\citep{Srinivas2020CURLCU}, and D4PG-Pixels~\citep{tassa2018deepmind}. CURL, DREAMER, and Pixel SAC are for reference only as they represent the state the art for low-data experiments (500K environment steps). These three are not perfectly comparable baselines; however, D4PG-Pixels is run on a comparable scale with 100 million environment steps. Because \agent{} relies on large scale distributed experience, we have a much larger number of available environment steps per gradient update. We run for 100M environment steps as with D4PG-Pixels, and we compute performance for our approaches by taking the evaluation performance of the final 10\% of steps. Across the majority of tasks, \agent{} outperforms D4PG-Pixels. The increase in performance is especially apparent for the more difficult tasks. For most of the easier tasks, the performance difference between the \agent{}, R2D2-GTrXL, and R2D2 is negligible. For  {\ttfamily ball\_in\_cup catch}, {\ttfamily cartpole swingup},  {\ttfamily finger spin} and {\ttfamily reacher easy}, even the original R2D2 agent performs on par with the D4PG-Pixels baseline. On more difficult tasks such as {\ttfamily fish swim}, and {\ttfamily swimmer swimmer6}, there is a very large, appreciable difference between \agent{}, R2D2, and R2D2-GTrXL. The combination of the LSTM and transformer specifically makes a large difference here especially compared to D4PG-Pixels. Interestingly, this architecture is also very important for situations where the R2D2-based approaches underperform. For {\ttfamily cheetah run} and {\ttfamily walker walk}, \agent{} dramatically narrows the performance gap between R2D2 and state of the art (in App.~\ref{app:learning-curves} we report learning curves for each game). Figure \ref{fig:auc-control} shows that also in this domain, we see a significant AUC improvement for \agent{} when compared to R2D2-gTrXL, t(24)=1.609, p=.03.  

\paragraph{DMLab-30.} 

\begin{figure*}
\centering
  \includegraphics[scale=0.45]{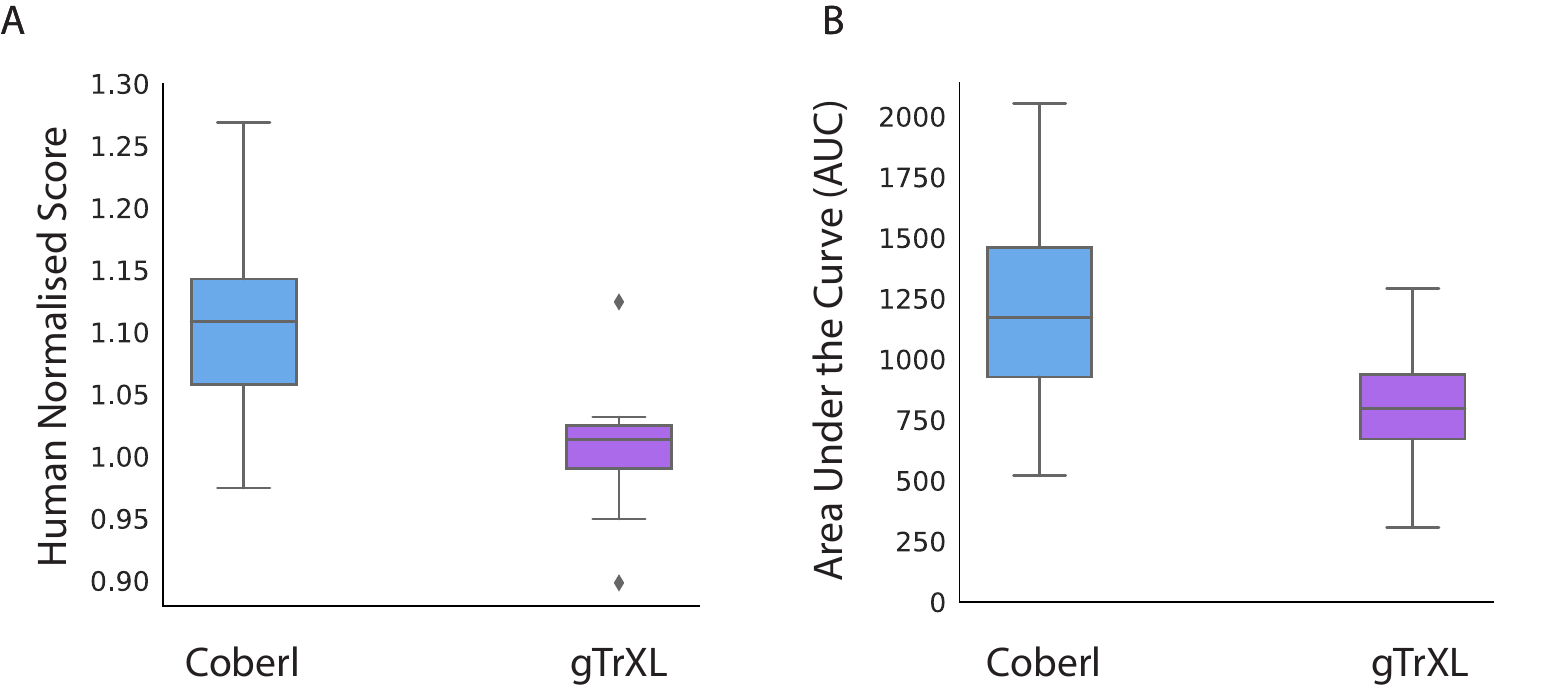}
  \vspace{-0.3cm}
  \caption{\small{DMLab-30 experiments. a) Human normalised returns in DMLab-30 across all the 30 levels (higher is better). b) Area under the curve for all the 30 DMLab levels. Results are over 5 seeds and the final 5\% of training.(lower) c) Area under the curve (AUC) for DMLab-30 across all the 30 levels. (higher is better)}} 
  \label{fig:dm_lab_plots}
  \vspace{-0.5cm}
\end{figure*}

To test \agent{} in a challenging 3 dimensional environment we run it in DmLab-30~\citep{beattie2016deepmind}. The agent was trained at the same time on all the 30 tasks, following the setup of GTrXl~\citep{parisotto2020stabilizing}, which we use as our baseline. In Figure~\ref{fig:dm_lab_plots}A we show the final results on the DMLab-30 domain. If we look at all the 30 games, \agent{} reaches a substantially higher score than GTrXL (\agent{}=115.47\% $\pm$ 4.21\%, GTrXL=101.54\% $\pm$ 0.50\%, t(60)=4.37, p=1.14e-5, Figure \ref{fig:dm_lab_plots}A).
In Figure~\ref{fig:dm_lab_plots}B we analysed data efficiency by computing the AUC and associated average statistics. In DMLab the difference between models is even more significant than Atari, t(60)=6.097, p=9.39e-09, with \agent{} (M=1193.28, SD=383.29) having better average AUC than GTrXL (M=764.09, SD=224.51), hence showing that our methods scales well to more complex domains. (see Appendix \ref{app:learning-curves} for learning curves)
% We also analysed the number of steps required to reach 100\% human normalised score, a measure for data efficiency. In this respect, \agent{} requires considerably fewer environment frames than GTrXL (\agent{}=2.96 $\pm$ 0.35 Billion, GTrXL=3.64 $\pm$ 0.43 Billion, see Figure~\ref{fig:dm_lab_plots}B).  Figure~\ref{fig:dm_lab_plots}C reports the AUC in DMLab, where the difference was even more significant than Atari, t(60)=5.869, p=2.23e-07, with \agent{} (M=1251.13, SD=366.86) having better average AUC than GTrXL (M=829.76, SD=229.09), hence showing that our methods scales well to more complex domains. (see Appendix \ref{app:learning-curves} for learning curves)

\subsection{Ablations}
\label{sec:ablations}

In Sec.~\ref{sec:method}, we explained contributions that are essential to \agent{}. We now disentangle the added benefit of these two separate contributions. Moreover, we run a set of ablations to understand the role of model size on the results. Ablations are run on 7 Atari games chosen to match the ones in the original DQN publication \citep{mnih2013playing}, and on all the 30 DMLab games.

\subsubsection{Impact of auxiliary losses}
\label{sec:aux-loss-ablations}

In Table~\ref{tab:ablation_aux_losses} we show that our contrastive loss contributes to a significant gain in performance, both in Atari and DMLab-30, when compared to \agent{} without it. Also, in challenging environments like DmLab-30, \agent{} without extra loss is still superior to the relative baseline. The only case where we do not see and advantage of using the auxiliary loss is if we consider the median score on the reduced ablation set of Atari games. However in the case of the DmLab-30, where we consider a larger set of levels (7 vs. 30), there is a clear benefit of the auxiliary loss. 

Moreover, Table~\ref{tab:ablation_extra_aux_losses} reports a comparison between our loss, SimCLR \citep{chen20icml} and CURL \citep{Srinivas2020CURLCU}. Although simpler than both SimCLR - which in its original implementation requires handcrafted augmentations - and CURL - which requires an additional network - our contrastive method shows improved performance. These experiments where run only on Atari to reduce computational costs while still being sufficient for the analysis. We also ran one extra ablation study where we did not use masking of the input to make our setup as close as possible to \citep{mitrovic2020representation}. As seen on Table~\ref{tab:ablation_extra_aux_losses}, the column CoBERL w/o masking shows that removing the masking technique derived from the language literature makes the results substantially worse.

\begin{table}[h]
% \resizebox{\textwidth}{!}{%
    \centering
    \footnotesize
    \begin{tabular}{c|c|l|p{2.7cm}|p{2.7cm}}
    \multicolumn{2}{c|}{}  & \agent{} & \agent{} w/o aux loss &  GTrXL  baseline* \\ \hline
    \multirow{2}{*}{DMLab-30} & Mean   & $\mathbf{115.47\%\pm4.21\%}$  & $105.29\%\pm2.02\%$  & $101.54\%\pm0.50\%$ \\ \cline{2-5} 
                          & Median & $\mathbf{110.86\%}$ & $104.84\%$ & $101.35\%$ \\ \hline
    \multirow{2}{*}{Atari} & Mean   & 
    $\mathbf{726.34\%\pm44.06\%}$  & $490.61\%\pm22.48\%$  & $605.62\%\pm47.46\%$ \\ \cline{2-5} 
    & Median & $270.03\%$ & $357.96\%$ & $\mathbf{394.53\%}$   \\ \hline
    \end{tabular}    
% }
\vspace{-0.2cm}
    \caption{\small{Impact of contrastive loss. Human normalized scores on Atari-57 ablation tasks and DMLab-30 tasks. * for DMLab the baseline is GTrXL trained with VMPO, for Atari the baseline is GTrXL trained with R2D2.}}
    \label{tab:ablation_aux_losses}

\end{table}

\vspace{-0.2cm}
\begin{table}[h]
\resizebox{\textwidth}{!}{%
    \centering
    \begin{tabular}{c|c|p{2.9cm}|p{2.9cm}|p{2.9cm}|p{2.9cm}}
    \multicolumn{2}{c|}{} & \agent{} & \agent{} with CURL &  \agent{} with SimCLR & \agent{} w/o masking \\ \hline
    \multirow{2}{*}{Atari} & Mean & $\mathbf{726.34\%\pm44.06\%}$  &  $608.53\%\pm62.93\%$ & $347.83\%\pm46.74\%$ & $331.05\%\pm59.05\%$\\ \cline{2-6} 
    & Median & $270.03\%$ & $\mathbf{272.34\%}$  & $265.00\%$ & $233.63\%$\\ \hline
    \end{tabular}    
}
\vspace{-0.4cm}
    \caption{\small{Comparison with alternative auxiliary losses.}}
    \label{tab:ablation_extra_aux_losses}
\vspace{-.5cm}
\end{table}

\subsubsection{Impact of architectural changes}
\label{sec:architectural-ablations}

Table~\ref{tab:ablation_architecture_results} shows the effects of removing the LSTM from \agent{} (column ``w/o LSTM''), as well as removing the gate and its associated skip connection (column ``w/o Gate''). In both cases \agent{} performs substantially worse showing that both components are needed.
Finally, we also experimented with substituting the learned gate with either a sum or a concatenation. The results, presented in Appendix~\ref{app:additional-results}, show that in most occasions these alternatives decrease performance, but not as substantially as removing the LSTM, gate or skip connections. Our hypothesis is that the learned gate gives more flexibility in complex environments, we leave it open for future work to explore this.

\begin{table}[ht!]
\centering
\footnotesize
% \resizebox{\textwidth}{!}{%
\begin{tabular}{c|c|l|p{2.7cm}|p{2.7cm}}
    \multicolumn{2}{c|}{} & \agent{} & w/o LSTM & w/o Gate\\ \hline
    \multirow{2}{*}{DMLab-30} & Mean   & $\mathbf{115.47\%\pm4.21\%}$  & $96.29\%\pm3.39$ & $86.42\%\pm7.25\%$  \\ \cline{2-5} 
    & Median & $\mathbf{110.86\%}$ & $99.83\%$ & $95.21\%$ \\ \hline
    \multirow{2}{*}{Atari} & Mean   & $\mathbf{726.34\%\pm44.06}$ & $504.42\%\pm48.51\%$ & $528.61\% \pm 73.21\%$   \\ \cline{2-5} 
    & Median & $270.03\%$ & $249.62\%$ & $\mathbf{283.16}\%$  \\ \hline
    \end{tabular}
    % }
    \vspace{-0.3cm}
    \caption{\small{Impact of Architectural changes. Human normalized scores on Atari ablation tasks and DMLab-30.}}
    \label{tab:ablation_architecture_results}
\end{table}

\subsubsection{Impact of number of parameters}

Table~\ref{tab:ablation_num_params} compares the models in terms of the number of parameters. For Atari, the number of parameters added by \agent{} over the R2D2(GTrXL) baseline is very limited; however, \agent{} still produces a significant gain in performance. We also tried to move the LSTM before the transformer module (column ``\agent{} with LSTM before''). In this case the representations for the contrastive loss were taken from before the LSTM. Interestingly, this setting performs worse, despite having the same number of parameters as \agent{}. This goes in the direction of our hypothesis that having the LSTM after the Transformer could allow the former to exploit the extra context provided by the latter. For DMLab-30, it is worth noting that \agent{} has a memory size of $256$, whereas GTrXL has a memory of size $512$ resulting in substantially fewer parameters. Nevertheless, the discrepancies between models are even more pronounced, even though the number of parameters is either exactly the same (``\agent{} with LSTM before'') or higher (GTrXL). This ablation is of particular interest as it shows that the results are driven by the particular architectural choice rather than the added parameters. Also it helps supporting our idea that the LSTM, by processing a certain amount of contextual information, allow for a reduced memory size on the Transformer, without a loss in performance and with a reduce computation complexity (given fewer parameters).

\begin{table}[ht!]
\centering
\resizebox{\textwidth}{!}{%
\begin{tabular}{c|c|p{2.7cm}|p{2.7cm}|p{2.7cm}|p{2.7cm}}
    \multicolumn{2}{c|}{} & \agent{} & GTrXL* & \agent{} with LSTM before & R2D2 LSTM \\ \hline
    \multirow{3}{*}{DMLab-30} & Mean   & $\mathbf{115.47\%\pm4.21\%}$  & $101.54\%\pm0.50\%$ & $101.29\%\pm1.80\%$ & N/A  \\ \cline{2-6} 
    & Median & $\mathbf{110.86\%}$ & $103.82\%$ & $100.03\%$ & N/A \\ \cline{2-6}
    & Num. Params. & $47$M & $66$M & $47$M & N/A \\ \hline
    \multirow{3}{*}{Atari} & Mean & $\mathbf{726.34\%\pm44.06}$ & $605.62\%\pm47.46\%$ & $604.36\%\pm78.60\%$ & $362.38\%\pm20.28\%$  \\ \cline{2-6} 
    & Median & $270.03\%$ & $\mathbf{394.53\%}$ & $296.72\%$ & $202.71\%$ \\ \cline{2-6}
    & Num. Params. & $46$M & $42$M & $46$M  & $18$M \\ \hline
    \end{tabular}
    }
    \vspace{-0.3cm}
    \caption{\small{Effect of number of parameters. Human normalized scores on Atari-57 ablation tasks and DMLab-30 tasks. *for DMLab-30 the baseline is GTrXL trained with VMPO with a memory size of $512$, for Atari the baseline is GTrXL trained with R2D2 with a memory size of $64$.}}
    \label{tab:ablation_num_params}
    \vspace{-0.5cm}
\end{table}

\section{Conclusions}
\label{sec:conclusions}

We proposed a novel RL agent, Contrastive BERT for RL (\agent{}), which introduces a new contrastive representation learning loss 
that enables the agent to efficiently learn consistent representations.  This, paired with an improved architecture, resulted in better data efficiency on a varied set of environments and tasks. 
% On Atari, \agent{} comfortably outperformed competing methods surpassing the human benchmark in 49 out of the 57 games. In the the DeepMind Control Suite, \agent{} showed significant improvement over previous work in 3 out of 6 tasks, while matching previous state-of-the-art in the remaining 3 tasks.
% On DMLab-30, \agent{} was significantly more data-efficient than GTrXl and also obtained higher final scores. 
Moreover, through an extensive set of ablation experiments we confirmed that all \agent{} components are necessary to achieve the performance of the final agent. 
Critically, \agent{} is fully end-to-end as it does not require an extra encoder network (vs CURL or \relic{}), or data augmentations (vs SIMCLR and \relic{}) and it has fewer parameters then gTrXL.
To conclude, we have shown that our auxiliary loss and architecture provide an effective and general means to efficiently train large attentional models in RL. (For extra conclusions and limitations see App.\ref{app:limitations})

\section{Optional reproducibility statement}
\label{sec:repro}

To help with reproducibility we have included extensive details for 
\begin{itemize}
    \item Setup details in Appendix~\ref{app:setup-details}
    \item Architecture description in Appendix~\ref{app:architecture}
    \item Hyper-parameters chosen in Appendix~\ref{app:hyperparameters}
\end{itemize}

We also report the pseudo-code for the algorithm and the auxiliary loss in Appendix~\ref{app:pseudo-code}

\clearpage

\bibliography{iclr2022_conference}

\begin{thebibliography}{52}
\providecommand{\natexlab}[1]{#1}
\providecommand{\url}[1]{\texttt{#1}}
\expandafter\ifx\csname urlstyle\endcsname\relax
  \providecommand{\doi}[1]{doi: #1}\else
  \providecommand{\doi}{doi: \begingroup \urlstyle{rm}\Url}\fi

\bibitem[Abdolmaleki et~al.(2018)Abdolmaleki, Springenberg, Degrave, Bohez,
  Tassa, Belov, Heess, and Riedmiller]{abdolmaleki2018relative}
Abbas Abdolmaleki, Jost~Tobias Springenberg, Jonas Degrave, Steven Bohez, Yuval
  Tassa, Dan Belov, Nicolas Heess, and Martin Riedmiller.
\newblock Relative entropy regularized policy iteration.
\newblock \emph{arXiv preprint arXiv:1812.02256}, 2018.

\bibitem[Agarwal et~al.(2021)Agarwal, Machado, Castro, and
  Bellemare]{Agarwal2021ContrastiveBS}
Rishabh Agarwal, Marlos~C. Machado, P.~S. Castro, and Marc~G. Bellemare.
\newblock Contrastive behavioral similarity embeddings for generalization in
  reinforcement learning.
\newblock \emph{ArXiv}, abs/2101.05265, 2021.

\bibitem[Beattie et~al.(2016)Beattie, Leibo, Teplyashin, Ward, Wainwright,
  K{\"u}ttler, Lefrancq, Green, Vald{\'e}s, Sadik, et~al.]{beattie2016deepmind}
Charles Beattie, Joel~Z Leibo, Denis Teplyashin, Tom Ward, Marcus Wainwright,
  Heinrich K{\"u}ttler, Andrew Lefrancq, Simon Green, V{\'\i}ctor Vald{\'e}s,
  Amir Sadik, et~al.
\newblock Deepmind lab.
\newblock \emph{arXiv preprint arXiv:1612.03801}, 2016.

\bibitem[Bellemare et~al.(2013)Bellemare, Naddaf, Veness, and
  Bowling]{bellemare2013arcade}
Marc~G Bellemare, Yavar Naddaf, Joel Veness, and Michael Bowling.
\newblock The arcade learning environment: An evaluation platform for general
  agents.
\newblock \emph{Journal of Artificial Intelligence Research}, 47:\penalty0
  253--279, 2013.

\bibitem[Brown et~al.(2020)Brown, Mann, Ryder, Subbiah, Kaplan, Dhariwal,
  Neelakantan, Shyam, Sastry, Askell, Agarwal, Herbert-Voss, Krueger, Henighan,
  Child, Ramesh, Ziegler, Wu, Winter, Hesse, Chen, Sigler, Litwin, Gray, Chess,
  Clark, Berner, McCandlish, Radford, Sutskever, and Amodei]{brown2020language}
Tom~B. Brown, Benjamin Mann, Nick Ryder, Melanie Subbiah, Jared Kaplan,
  Prafulla Dhariwal, Arvind Neelakantan, Pranav Shyam, Girish Sastry, Amanda
  Askell, Sandhini Agarwal, Ariel Herbert-Voss, Gretchen Krueger, Tom Henighan,
  Rewon Child, Aditya Ramesh, Daniel~M. Ziegler, Jeffrey Wu, Clemens Winter,
  Christopher Hesse, Mark Chen, Eric Sigler, Mateusz Litwin, Scott Gray,
  Benjamin Chess, Jack Clark, Christopher Berner, Sam McCandlish, Alec Radford,
  Ilya Sutskever, and Dario Amodei.
\newblock Language models are few-shot learners.
\newblock In \emph{NeurIPS}, 2020.

\bibitem[Carion et~al.(2020)Carion, Massa, Synnaeve, Usunier, Kirillov, and
  Zagoruyko]{carion2020endtoend}
Nicolas Carion, Francisco Massa, Gabriel Synnaeve, Nicolas Usunier, Alexander
  Kirillov, and Sergey Zagoruyko.
\newblock End-to-end object detection with transformers.
\newblock In \emph{ECCV}, 2020.

\bibitem[Caron et~al.(2020)Caron, Misra, Mairal, Goyal, Bojanowski, and
  Joulin]{Caron2020UnsupervisedLO}
M.~Caron, I.~Misra, J.~Mairal, Priya Goyal, P.~Bojanowski, and Armand Joulin.
\newblock Unsupervised learning of visual features by contrasting cluster
  assignments.
\newblock \emph{ArXiv}, abs/2006.09882, 2020.

\bibitem[Chen et~al.(2020)Chen, Kornblith, Norouzi, and Hinton]{chen20icml}
Ting Chen, Simon Kornblith, Mohammad Norouzi, and Geoffrey Hinton.
\newblock A simple framework for contrastive learning of visual
  representations.
\newblock In \emph{Proceedings of the 37th International Conference on Machine
  Learning}, pp.\  1597--1607, 2020.

\bibitem[Dai et~al.(2019)Dai, Yang, Yang, Carbonell, Le, and
  Salakhutdinov]{dai2019transformer}
Zihang Dai, Zhilin Yang, Yiming Yang, Jaime Carbonell, Quoc~V Le, and Ruslan
  Salakhutdinov.
\newblock Transformer-xl: Attentive language models beyond a fixed-length
  context.
\newblock \emph{arXiv preprint arXiv:1901.02860}, 2019.

\bibitem[Dehghani et~al.(2018)Dehghani, Gouws, Vinyals, Uszkoreit, and
  Kaiser]{dehghani2018universal}
Mostafa Dehghani, Stephan Gouws, Oriol Vinyals, Jakob Uszkoreit, and {\L}ukasz
  Kaiser.
\newblock Universal transformers.
\newblock \emph{arXiv preprint arXiv:1807.03819}, 2018.

\bibitem[Devlin et~al.(2019)Devlin, Chang, Lee, and
  Toutanova]{devlinetal2019bert}
Jacob Devlin, Ming-Wei Chang, Kenton Lee, and Kristina Toutanova.
\newblock {BERT}: Pre-training of deep bidirectional transformers for language
  understanding.
\newblock In \emph{Proceedings of the 2019 Conference of the North {A}merican
  Chapter of the Association for Computational Linguistics: Human Language
  Technologies, Volume 1 (Long and Short Papers)}, pp.\  4171--4186, 2019.

\bibitem[Dosovitskiy et~al.(2021)Dosovitskiy, Beyer, Kolesnikov, Weissenborn,
  Zhai, Unterthiner, Dehghani, Minderer, Heigold, Gelly, Uszkoreit, and
  Houlsby]{dosovitskiy2021an}
Alexey Dosovitskiy, Lucas Beyer, Alexander Kolesnikov, Dirk Weissenborn,
  Xiaohua Zhai, Thomas Unterthiner, Mostafa Dehghani, Matthias Minderer, Georg
  Heigold, Sylvain Gelly, Jakob Uszkoreit, and Neil Houlsby.
\newblock An image is worth 16x16 words: Transformers for image recognition at
  scale.
\newblock In \emph{International Conference on Learning Representations}, 2021.

\bibitem[Fortunato et~al.(2019)Fortunato, Tan, Faulkner, Hansen, Badia,
  Buttimore, Deck, Leibo, and Blundell]{fortunato2019generalization}
Meire Fortunato, Melissa Tan, Ryan Faulkner, Steven Hansen,
  Adri{\`a}~Puigdom{\`e}nech Badia, Gavin Buttimore, Charlie Deck, Joel~Z
  Leibo, and Charles Blundell.
\newblock Generalization of reinforcement learners with working and episodic
  memory.
\newblock \emph{NeurIPS}, 2019.

\bibitem[Gutmann \& Hyv{\"a}rinen(2010)Gutmann and
  Hyv{\"a}rinen]{gutmann2010noise}
Michael Gutmann and Aapo Hyv{\"a}rinen.
\newblock Noise-contrastive estimation: A new estimation principle for
  unnormalized statistical models.
\newblock In \emph{Proceedings of the Thirteenth International Conference on
  Artificial Intelligence and Statistics}, pp.\  297--304, 2010.

\bibitem[Ha \& Schmidhuber(2018)Ha and Schmidhuber]{ha2018world}
David Ha and J{\"u}rgen Schmidhuber.
\newblock World models.
\newblock \emph{arXiv preprint arXiv:1803.10122}, 2018.

\bibitem[Haarnoja et~al.(2018)Haarnoja, Zhou, Abbeel, and
  Levine]{HaarnojaZAL18}
Tuomas Haarnoja, Aurick Zhou, Pieter Abbeel, and Sergey Levine.
\newblock Soft actor-critic: Off-policy maximum entropy deep reinforcement
  learning with a stochastic actor.
\newblock In \emph{ICML}, 2018.

\bibitem[Hadsell et~al.(2006)Hadsell, Chopra, and
  LeCun]{hadsell2006dimensionality}
Raia Hadsell, Sumit Chopra, and Yann LeCun.
\newblock Dimensionality reduction by learning an invariant mapping.
\newblock In \emph{2006 IEEE Computer Society Conference on Computer Vision and
  Pattern Recognition (CVPR'06)}, volume~2, pp.\  1735--1742. IEEE, 2006.

\bibitem[Hafner et~al.(2020{\natexlab{a}})Hafner, Lillicrap, Ba, and
  Norouzi]{Hafner2020Dream}
Danijar Hafner, Timothy Lillicrap, Jimmy Ba, and Mohammad Norouzi.
\newblock Dream to control: Learning behaviors by latent imagination.
\newblock In \emph{ICLR}, 2020{\natexlab{a}}.

\bibitem[Hafner et~al.(2020{\natexlab{b}})Hafner, Lillicrap, Norouzi, and
  Ba]{hafner2020mastering}
Danijar Hafner, Timothy Lillicrap, Mohammad Norouzi, and Jimmy Ba.
\newblock Mastering atari with discrete world models.
\newblock \emph{arXiv preprint arXiv:2010.02193}, 2020{\natexlab{b}}.

\bibitem[Hessel et~al.(2018)Hessel, Modayil, Van~Hasselt, Schaul, Ostrovski,
  Dabney, Horgan, Piot, Azar, and Silver]{hessel2018rainbow}
Matteo Hessel, Joseph Modayil, Hado Van~Hasselt, Tom Schaul, Georg Ostrovski,
  Will Dabney, Dan Horgan, Bilal Piot, Mohammad Azar, and David Silver.
\newblock Rainbow: Combining improvements in deep reinforcement learning.
\newblock In \emph{Proceedings of the AAAI Conference on Artificial
  Intelligence}, volume~32, 2018.

\bibitem[Hessel et~al.(2021)Hessel, Kroiss, Clark, Kemaev, Quan, Keck, Viola,
  and Hasselt]{hessel2021podracer}
Matteo Hessel, Manuel Kroiss, Aidan Clark, Iurii Kemaev, John Quan, Thomas
  Keck, Fabio Viola, and Hado~van Hasselt.
\newblock Podracer architectures for scalable reinforcement learning, 2021.

\bibitem[Hochreiter \& Schmidhuber(1997)Hochreiter and
  Schmidhuber]{hochreiter1997long}
Sepp Hochreiter and J{\"u}rgen Schmidhuber.
\newblock Long short-term memory.
\newblock \emph{Neural computation}, 9\penalty0 (8):\penalty0 1735--1780, 1997.

\bibitem[Jaderberg et~al.(2016)Jaderberg, Mnih, Czarnecki, Schaul, Leibo,
  Silver, and Kavukcuoglu]{jaderberg2016reinforcement}
Max Jaderberg, Volodymyr Mnih, Wojciech~Marian Czarnecki, Tom Schaul, Joel~Z
  Leibo, David Silver, and Koray Kavukcuoglu.
\newblock Reinforcement learning with unsupervised auxiliary tasks.
\newblock \emph{arXiv preprint arXiv:1611.05397}, 2016.

\bibitem[Kaiser et~al.(2019)Kaiser, Babaeizadeh, Milos, Osinski, Campbell,
  Czechowski, Erhan, Finn, Kozakowski, Levine, et~al.]{kaiser2019model}
Lukasz Kaiser, Mohammad Babaeizadeh, Piotr Milos, Blazej Osinski, Roy~H
  Campbell, Konrad Czechowski, Dumitru Erhan, Chelsea Finn, Piotr Kozakowski,
  Sergey Levine, et~al.
\newblock Model-based reinforcement learning for atari.
\newblock \emph{arXiv preprint arXiv:1903.00374}, 2019.

\bibitem[Kapturowski et~al.(2018)Kapturowski, Ostrovski, Quan, Munos, and
  Dabney]{kapturowski2018recurrent}
Steven Kapturowski, Georg Ostrovski, John Quan, Remi Munos, and Will Dabney.
\newblock Recurrent experience replay in distributed reinforcement learning.
\newblock In \emph{International conference on learning representations}, 2018.

\bibitem[Kostrikov et~al.(2020)Kostrikov, Yarats, and
  Fergus]{kostrikov2020image}
Ilya Kostrikov, Denis Yarats, and Rob Fergus.
\newblock Image augmentation is all you need: Regularizing deep reinforcement
  learning from pixels.
\newblock \emph{arXiv preprint arXiv:2004.13649}, 2020.

\bibitem[Lake et~al.(2017)Lake, Ullman, Tenenbaum, and
  Gershman]{lake2017building}
Brenden~M Lake, Tomer~D Ullman, Joshua~B Tenenbaum, and Samuel~J Gershman.
\newblock Building machines that learn and think like people.
\newblock \emph{Behavioral and brain sciences}, 40, 2017.

\bibitem[Liu et~al.(2021)Liu, Zhang, Zhao, Qin, Zhu, Jian, Yu, and
  Liu]{liu2021returnbased}
Guoqing Liu, Chuheng Zhang, Li~Zhao, Tao Qin, Jinhua Zhu, Li~Jian, Nenghai Yu,
  and Tie-Yan Liu.
\newblock Return-based contrastive representation learning for reinforcement
  learning.
\newblock In \emph{ICLR 2021}, January 2021.

\bibitem[Machado et~al.(2018)Machado, Bellemare, Talvitie, Veness, Hausknecht,
  and Bowling]{machado2018revisiting}
Marlos~C Machado, Marc~G Bellemare, Erik Talvitie, Joel Veness, Matthew
  Hausknecht, and Michael Bowling.
\newblock Revisiting the arcade learning environment: Evaluation protocols and
  open problems for general agents.
\newblock \emph{Journal of Artificial Intelligence Research}, 61:\penalty0
  523--562, 2018.

\bibitem[Mazoure et~al.(2020)Mazoure, des Combes, Doan, Bachman, and
  Hjelm]{Mazoure2020DeepRA}
Bogdan Mazoure, R'emi~Tachet des Combes, Thang Doan, Philip Bachman, and
  R.~Devon Hjelm.
\newblock Deep reinforcement and infomax learning.
\newblock \emph{ArXiv}, abs/2006.07217, 2020.

\bibitem[McCandlish et~al.(2018)McCandlish, Kaplan, Amodei, and
  Team]{mccandlish2018empirical}
Sam McCandlish, Jared Kaplan, Dario Amodei, and OpenAI~Dota Team.
\newblock An empirical model of large-batch training.
\newblock \emph{arXiv preprint arXiv:1812.06162}, 2018.

\bibitem[Mitrovic et~al.(2021)Mitrovic, McWilliams, Walker, Buesing, and
  Blundell]{mitrovic2020representation}
Jovana Mitrovic, Brian McWilliams, Jacob Walker, Lars Buesing, and Charles
  Blundell.
\newblock Representation learning via invariant causal mechanisms.
\newblock In \emph{International conference on learning representations}, 2021.

\bibitem[Mnih et~al.(2013)Mnih, Kavukcuoglu, Silver, Graves, Antonoglou,
  Wierstra, and Riedmiller]{mnih2013playing}
Volodymyr Mnih, Koray Kavukcuoglu, David Silver, Alex Graves, Ioannis
  Antonoglou, Daan Wierstra, and Martin Riedmiller.
\newblock Playing atari with deep reinforcement learning.
\newblock \emph{arXiv preprint arXiv:1312.5602}, 2013.

\bibitem[Mnih et~al.(2015)Mnih, Kavukcuoglu, Silver, Rusu, Veness, Bellemare,
  Graves, Riedmiller, Fidjeland, Ostrovski, et~al.]{mnih2015human}
Volodymyr Mnih, Koray Kavukcuoglu, David Silver, Andrei~A Rusu, Joel Veness,
  Marc~G Bellemare, Alex Graves, Martin Riedmiller, Andreas~K Fidjeland, Georg
  Ostrovski, et~al.
\newblock Human-level control through deep reinforcement learning.
\newblock \emph{nature}, 518\penalty0 (7540):\penalty0 529--533, 2015.

\bibitem[Oord et~al.(2018)Oord, Li, and Vinyals]{oord2018representation}
Aaron van~den Oord, Yazhe Li, and Oriol Vinyals.
\newblock Representation learning with contrastive predictive coding.
\newblock \emph{arXiv preprint arXiv:1807.03748}, 2018.

\bibitem[Parisotto et~al.(2020)Parisotto, Song, Rae, Pascanu, Gulcehre,
  Jayakumar, Jaderberg, Kaufman, Clark, Noury,
  et~al.]{parisotto2020stabilizing}
Emilio Parisotto, Francis Song, Jack Rae, Razvan Pascanu, Caglar Gulcehre,
  Siddhant Jayakumar, Max Jaderberg, Raphael~Lopez Kaufman, Aidan Clark, Seb
  Noury, et~al.
\newblock Stabilizing transformers for reinforcement learning.
\newblock In \emph{International Conference on Machine Learning}, pp.\
  7487--7498. PMLR, 2020.

\bibitem[Peng \& Williams(1994)Peng and Williams]{peng1994incremental}
Jing Peng and Ronald~J Williams.
\newblock Incremental multi-step q-learning.
\newblock In \emph{Machine Learning Proceedings 1994}, pp.\  226--232.
  Elsevier, 1994.

\bibitem[Schmidhuber(1990)]{schmidhuber1990making}
Jiirgen Schmidhuber.
\newblock Making the world differentiable: On using self-supervised fully
  recurrent n eu al networks for dynamic reinforcement learning and planning in
  non-stationary environm nts.
\newblock 1990.

\bibitem[Schrittwieser et~al.(2020)Schrittwieser, Antonoglou, Hubert, Simonyan,
  Sifre, Schmitt, Guez, Lockhart, Hassabis, Graepel,
  et~al.]{schrittwieser2020mastering}
Julian Schrittwieser, Ioannis Antonoglou, Thomas Hubert, Karen Simonyan,
  Laurent Sifre, Simon Schmitt, Arthur Guez, Edward Lockhart, Demis Hassabis,
  Thore Graepel, et~al.
\newblock Mastering atari, go, chess and shogi by planning with a learned
  model.
\newblock \emph{Nature}, 588\penalty0 (7839):\penalty0 604--609, 2020.

\bibitem[Schwarzer et~al.(2020)Schwarzer, Anand, Goel, Hjelm, Courville, and
  Bachman]{Schwarzer2020DataEfficientRL}
Max Schwarzer, Ankesh Anand, R.~Goel, R.~Devon Hjelm, Aaron~C. Courville, and
  Philip Bachman.
\newblock Data-efficient reinforcement learning with momentum predictive
  representations.
\newblock \emph{ArXiv}, abs/2007.05929, 2020.

\bibitem[Schwarzschild et~al.(2021)Schwarzschild, Gupta, Ghiasi, Goldblum, and
  Goldstein]{schwarzschild2021uncanny}
Avi Schwarzschild, Arjun Gupta, Amin Ghiasi, Micah Goldblum, and Tom Goldstein.
\newblock The uncanny similarity of recurrence and depth.
\newblock \emph{arXiv preprint arXiv:2102.11011}, 2021.

\bibitem[Song et~al.(2019)Song, Abdolmaleki, Springenberg, Clark, Soyer, Rae,
  Noury, Ahuja, Liu, Tirumala, et~al.]{song2019v}
H~Francis Song, Abbas Abdolmaleki, Jost~Tobias Springenberg, Aidan Clark,
  Hubert Soyer, Jack~W Rae, Seb Noury, Arun Ahuja, Siqi Liu, Dhruva Tirumala,
  et~al.
\newblock V-mpo: On-policy maximum a posteriori policy optimization for
  discrete and continuous control.
\newblock \emph{arXiv preprint arXiv:1909.12238}, 2019.

\bibitem[Srinivas et~al.(2020{\natexlab{a}})Srinivas, Laskin, and
  Abbeel]{Srinivas2020CURLCU}
A.~Srinivas, M.~Laskin, and P.~Abbeel.
\newblock Curl: Contrastive unsupervised representations for reinforcement
  learning.
\newblock In \emph{ICML}, 2020{\natexlab{a}}.

\bibitem[Srinivas et~al.(2020{\natexlab{b}})Srinivas, Laskin, and
  Abbeel]{srinivas2020curl}
Aravind Srinivas, Michael Laskin, and Pieter Abbeel.
\newblock Curl: Contrastive unsupervised representations for reinforcement
  learning.
\newblock \emph{arXiv preprint arXiv:2004.04136}, 2020{\natexlab{b}}.

\bibitem[Sutton(1990)]{sutton1990integrated}
Richard~S Sutton.
\newblock Integrated architectures for learning, planning, and reacting based
  on approximating dynamic programming.
\newblock In \emph{Machine learning proceedings 1990}, pp.\  216--224.
  Elsevier, 1990.

\bibitem[Tassa et~al.(2018)Tassa, Doron, Muldal, Erez, Li, Casas, Budden,
  Abdolmaleki, Merel, Lefrancq, et~al.]{tassa2018deepmind}
Yuval Tassa, Yotam Doron, Alistair Muldal, Tom Erez, Yazhe Li, Diego de~Las
  Casas, David Budden, Abbas Abdolmaleki, Josh Merel, Andrew Lefrancq, et~al.
\newblock Deepmind control suite.
\newblock \emph{arXiv preprint arXiv:1801.00690}, 2018.

\bibitem[van Hasselt et~al.(2016)van Hasselt, Guez, Hessel, Mnih, and
  Silver]{van2016learning}
Hado van Hasselt, Arthur Guez, Matteo Hessel, Volodymyr Mnih, and David Silver.
\newblock Learning values across many orders of magnitude.
\newblock \emph{arXiv preprint arXiv:1602.07714}, 2016.

\bibitem[Vaswani et~al.(2017)Vaswani, Shazeer, Parmar, Uszkoreit, Jones, Gomez,
  Kaiser, and Polosukhin]{vaswani2017attention}
Ashish Vaswani, Noam Shazeer, Niki Parmar, Jakob Uszkoreit, Llion Jones,
  Aidan~N Gomez, Lukasz Kaiser, and Illia Polosukhin.
\newblock Attention is all you need.
\newblock In \emph{NeurIPS}, 2017.

\bibitem[Virtanen et~al.(2020)Virtanen, Gommers, Oliphant, Haberland, Reddy,
  Cournapeau, Burovski, Peterson, Weckesser, Bright, {van der Walt}, Brett,
  Wilson, Millman, Mayorov, Nelson, Jones, Kern, Larson, Carey, Polat, Feng,
  Moore, {VanderPlas}, Laxalde, Perktold, Cimrman, Henriksen, Quintero, Harris,
  Archibald, Ribeiro, Pedregosa, {van Mulbregt}, and {SciPy 1.0
  Contributors}]{2020SciPy-NMeth}
Pauli Virtanen, Ralf Gommers, Travis~E. Oliphant, Matt Haberland, Tyler Reddy,
  David Cournapeau, Evgeni Burovski, Pearu Peterson, Warren Weckesser, Jonathan
  Bright, St{\'e}fan~J. {van der Walt}, Matthew Brett, Joshua Wilson, K.~Jarrod
  Millman, Nikolay Mayorov, Andrew R.~J. Nelson, Eric Jones, Robert Kern, Eric
  Larson, C~J Carey, {\.I}lhan Polat, Yu~Feng, Eric~W. Moore, Jake
  {VanderPlas}, Denis Laxalde, Josef Perktold, Robert Cimrman, Ian Henriksen,
  E.~A. Quintero, Charles~R. Harris, Anne~M. Archibald, Ant{\^o}nio~H. Ribeiro,
  Fabian Pedregosa, Paul {van Mulbregt}, and {SciPy 1.0 Contributors}.
\newblock {{SciPy} 1.0: Fundamental Algorithms for Scientific Computing in
  Python}.
\newblock \emph{Nature Methods}, 17:\penalty0 261--272, 2020.
\newblock \doi{10.1038/s41592-019-0686-2}.

\bibitem[Voita et~al.(2019)Voita, Sennrich, and Titov]{voita2019bottom}
Elena Voita, Rico Sennrich, and Ivan Titov.
\newblock The bottom-up evolution of representations in the transformer: A
  study with machine translation and language modeling objectives.
\newblock \emph{arXiv preprint arXiv:1909.01380}, 2019.

\bibitem[Yang et~al.(2019)Yang, Dai, Yang, Carbonell, Salakhutdinov, and
  Le]{yang2019xlnet}
Zhilin Yang, Zihang Dai, Yiming Yang, Jaime Carbonell, Ruslan Salakhutdinov,
  and Quoc~V Le.
\newblock Xlnet: Generalized autoregressive pretraining for language
  understanding.
\newblock \emph{arXiv preprint arXiv:1906.08237}, 2019.

\bibitem[Zhu et~al.(2020)Zhu, Xia, Wu, Deng, Zhou, Qin, and
  Li]{Zhu2020MaskedCR}
Jinhua Zhu, Yingce Xia, Lijun Wu, Jiajun Deng, W.~Zhou, Tao Qin, and H.~Li.
\newblock Masked contrastive representation learning for reinforcement
  learning.
\newblock \emph{ArXiv}, abs/2010.07470, 2020.

\end{thebibliography}
\bibliographystyle{iclr2022_conference}
\clearpage

\appendix
\section{Setup details}
\label{app:setup-details}

\subsection{R2D2 Distributed system setup}
\label{app:distributed_R2D2}
Following R2D2, the distributed system consists of several parts: actors, a replay buffer, a learner, and an evaluator. Additionally, we introduce a centralized batched inference process to make more efficient use of actor resources.

{\bf Actors}: We use $512$ processes to interact with independent copies of the environment, called actors. They send the following information to a central batch inference process: 
\begin{itemize}
    \item $x_t$: the observation at time $t$.
    \item $r_{t-1}$: the reward at the previous time, initialized with $r_{-1}=0$.
    \item $a_{t-1}$: the action at the previous time, $a_{-1}$ is initialized to $0$.
    \item $h_{t-1}$: recurrent state at the previous time, is initialized with $h_{-1}=0$.
\end{itemize}

They block until they receive $Q(x_t, a; \theta)$. The $l$-th actor picks $a_t$ using an $\epsilon_l$-greedy policy. As R2D2, the value of $\epsilon_l$ is computed following:

\begin{equation*}
   \epsilon_l = \epsilon^{1+\alpha\frac{l}{L-1}} 
\end{equation*}  

where $\epsilon=0.4$ and  $\alpha=7$. After that is computed, the actors send the experienced transition information to the replay buffer.

{\bf Batch inference process}: This central batch inference process receives the inputs mentioned above from all actors. This process has the same architecture as the learner with weights that are fetched from the learner every $0.5$ seconds. The process blocks until a sufficient amount of actors have sent inputs, forming a batch. We use a batch size of $64$ in our experiments. After a batch is formed, the neural network of the agent is run to compute $Q(x_t, a, \theta)$ for the whole batch, and these values are sent to their corresponding actors.

{\bf Replay buffer:} it stores fixed-length sequences of \emph{transitions} $T=(\omega_s)_{s=t}^{t+L-1}$ along with their priorities $p_T$, where $L$ is the trace length we use.
A transition is of the form $\omega_s=(r_{s-1}, a_{s-1}, h_{s-1}, x_s, a_s, h_s, r_{s}, x_{s+1})$. Concretely, this consists of the following elements:
\begin{itemize}
    \item $r_{s-1}$: reward at the previous time.
    \item $a_{s-1}$: action done by the agent at the previous time.
    \item $h_{s-1}$: recurrent state (in our case hidden state of the LSTM) at the previous time.
    \item $x_s$: observation provided by the environment at the current time.
    \item $a_s$: action done by the agent at the current time.
    \item $h_{s}$: recurrent state (in our case hidden state of the LSTM) at the current time.
    \item $r_s$: reward at the current time.
    \item $x_{s+1}$: observation provided by the environment at the next time.
\end{itemize}

The sequences never cross episode boundaries and they are stored into the buffer in an overlapping fashion, by an amount which we call the \emph{replay period}. Finally, concerning the priorities, we followed the same prioritization scheme proposed by~\cite{kapturowski2018recurrent} using a mixture of max and mean of the TD-errors in the sequence with priority exponent $\eta= 0.9$.

{\bf Evaluator}: the evaluator shares the same network architecture as the learner, with weights that are fetched from the learner every episode. Unlike the actors, the experience produced by the evaluator is not sent to the replay buffer. The evaluator acts in the same way as the actors, except that all the computation is done within the single CPU process instead of delegating inference to the batch inference process. At the end of $5$ episodes the results of those $5$ episodes are average and reported. In this paper we report the average performance provided by such reports over the last $5\%$ frames (for example, on Atari this is the average of all the performance reports obtained when the total frames consumed by actors is between 190M and 200M frames).

{\bf Learner}: The learner contains two identical networks called the online and target networks with different weights $\theta$ and $\theta'$ respectively~\citep{mnih2015human}.
The target network's weights $\theta'$ are updated to $\theta$ every $400$ optimization steps. $\theta$ is updated by executing the following sequence of instructions:
\begin{itemize}
    \item First, the learner samples a batch of size $64$ (batch size) of fixed-length sequences of transitions from the replay buffer, with each transition being of length $L$: $T_i=(\omega^i_s)_{s=t}^{t+L-1}$.
    \item Then, a forward pass is done on the online network and the target with inputs $(x^i_s, r^i_{s-1}, a^i_{s-1}, h^i_{s-1})_{s=t}^{t+H}$ in order to obtain the state-action values $\{(Q(x^i_s, a; \theta), Q(x^i_s, a; \theta')\}$.
    \item With $\{(Q(x^i_s, a; \theta), Q(x^i_s, a; \theta')\}$, the $Q(\lambda)$ loss is computed.
    \item The online network is used again to compute the auxiliary contrastive loss.
    \item Both losses are summed (with by weighting the auxiliary loss by $0.1$ as described in \ref{app:hyperparameters}), and optimized with an Adam optimizer.
    \item Finally, the priorities are computed for the sampled sequence of transitions and updated in the replay buffer.
\end{itemize}

\subsection{V-MPO distributed setup}
\label{app:distributed_VMPO}
For on-policy training, we used a Podracer setup similar to \citep{hessel2021podracer} for fast usage of experience from actors by learners. 

{\bf TPU learning and acting}: As in the Sebulba setup of \citep{hessel2021podracer}, acting and learning network computations were co-located on a set of TPU chips, split into a ratio of 3 cores used for learning for every 1 core used for inference. This ratio then scales with the total number of chips used.

{\bf Environment execution}: Due to the size of the recurrent states used by \agent{} and stored on the host CPU, it was not possible to execute the environments locally. To proceed we used 64 remote environment servers which serve only to step multiple copies of the environment. 1024 concurrent episodes were processed to balance frames per second, latency between acting and learning, and memory usage of the agent states on the host CPUs.

\subsection{Computation used}
\paragraph{R2D2}
We train the agent with a single TPU v2-based learner, performing approximately $5$ network updates per second (each update on a mini-batch of $64$ sequences of length $80$ for Atari and $120$ for Control). We use $512$ actors, using $4$ actors per CPU core, with each one performing $\sim 64$ environment steps per second on Atari. Finally for the batch inference process a TPU v2, which allows all actors to achieve the speed we have described. In particular, we used 8 TPU cores for learning and 2 for inference.

\paragraph{V-MPO}
We train the agent with 4 hosts each with 8 TPU v2 cores. Each of the 8 cores per host was split into 6 for learning and 2 for inference. We separately used 64 remote CPU environment servers to step 1024 concurrent environment episodes using the actions returned from inference. The learner updates were made up of a mini-batch of 120 sequences, each of length 95 frames. This setup enabled 4.6 network updates per second, or 53.4k frames per second.

\subsection{Complexity analysis}
As stated, the agent consists of layers of convolutions, transformer layers, and linear layers. Therefore the complexity is $max\{O(n^2\cdot d), O(k\cdot n\cdot d^2)\}$, where $k$ is the kernel size in the case of convolutions, $n$ is the size of trajectories, and $d$ is the size of hidden layers.

\clearpage
\section{Architecture description}
\label{app:architecture}
\subsection{Encoder}
As shown in Fig.~\ref{fig:arch}, observations $O_i$ are encoded using an encoder. In this work, the encoder we have used is a ResNet-47 encoder. Those 47 layers are divided in $4$ groups which have the following characteristics:

\begin{itemize}
    \item An initial stride-2 convolution with filter size $3$x$3$ ($1\cdot 4$ layers).
    \item Number of residual bottleneck blocks (in order): $(2, 4, 6, 2)$. Each block has 3 convolutional layers with ReLU activations, with filter sizes $1$x$1$, $3$x$3$, and $1$x$1$ respectively ($(2+4+6+2) \cdot 3$ layers).
    \item Number of channels for the last convolution in each block: $(64, 128, 256, 512)$.
    \item Number of channels for the non-last convolutions in each block: $(16, 32, 64, 128)$.
    \item Group norm is applied after each group, with a group size of $8$.
\end{itemize}

After this observation encoding step, a final 2-layer MLP with ReLU activations of sizes $(512, 448)$ is applied. The previous reward and one-hot encoded action are concatenated and projected with a linear layer into a $64$-dimensional vector. This $64$-dimensional vector is concatenated with the $448$-dimensional encoded input to have a final $512$-dimensional output.

\subsection{Transformer}
As described in Section~\ref{sec:method}, the output of the encoder is fed to a Gated Transformer XL. For Atari and Control, the transformer has the following characteristics:

\begin{itemize}
    \item Number of layers: $8$.
    \item Memory size: $64$.
    \item Hidden dimensions: $512$.
    \item Number of heads: $8$.
    \item Attention size: $64$.
    \item Output size: $512$.
    \item Activation function: GeLU.
\end{itemize}

For DmLab the transformer has the following characteristics:

\begin{itemize}
    \item Number of layers: $12$.
    \item Memory size: $256$ for \agent{} and $512$ for gTrXL.
    \item Hidden dimensions: $128$.
    \item Number of heads: $4$.
    \item Attention size: $64$.
    \item Output size: $512$.
    \item Activation function: ReLU.
\end{itemize}

the GTrXL baseline is identical, but with a Memory size of $512$.

\subsection{LSTM and Value head}

For both R2D2 and V-MPO the outputs of the transformer and encoder are passed through a GRU transform to obtain a $512$-dimensional vector. After that, an LSTM with $512$ hidden units is applied.
The the value function is estimated differently depending on the RL algorithm used.

\paragraph{R2D2} Following the LSTM, a Linear layer of size $512$ is used, followed by a ReLU activation. Finally, to compute the Q values from that $512$ vector a dueling head is used, as in~\cite{kapturowski2018recurrent}, a dueling head is is used, which requires a linear projection to the number of actions of the task, and another projection to a unidimensional vector.
\paragraph{V-MPO} Following the LSTM, a 2 layer MLP with size $512$ and $30$ (i.e. the number of levels in DMLab) is used. In the MLP we use ReLU activation. As we are interested in the multi-task setting where a single agent learns a large number of tasks with differing reward scales, we used PopArt \citep{van2016learning} for the value function estimation (see Table. \ref{tab:hyperparameters_VMPO} for details).

\subsection{Critic Function}
For DmLab-30 (V-MPO), we used a 2 layer MLP with hidden sizes 512 and 128. For Atari and Control Suite (R2D2) we used a single layer of size 512.

\clearpage
\section{Hyperparameters}
\label{app:hyperparameters}

For the experiments in Atari57 and the DeepMind Control suite, \agent{} uses the R2D2 distributed setup. We use $512$ actors for all our experiments. We do not constrain the amount of replay done for each experience trajectory that actors deposit in the buffer. However, we have found empirical replay frequency per data point to be close among all our experiments (with an expected value of $1.5$ samples per data point). We use a separate evaluator process that shares weights with our learner in order to measure the performance of our agents. We report scores at the end of training. The hyperparameters and architecture we choose for these two domains are the same with two exceptions: i) we use a shorter trace length for Atari ($80$ instead of $120$) as the environment does not require a long context to inform decisions, and ii) we use a squashing function on Atari and the Control Suite to transform our $Q$ values (as done in \citep{kapturowski2018recurrent}) since reward structures vary highly in magnitude between tasks.

\subsection{Atari and DMLab pre-processing}
\label{atari_hypers}
We use the commonly used input pre-processing on Atari and DMLab frames, shown on Tab.~\ref{table_hyper_atari}. One difference with the original work of~\cite{mnih2015human}, is that we do not use frame stacking, as we rely on our memory systems to be able to integrate information from the past, as done in \cite{kapturowski2018recurrent}. ALE is publicly available at \url{https://github.com/mgbellemare/Arcade-Learning-Environment}.

\begin{table}[h]
\centering
\begin{tabular}{l|c}
\textbf{Hyperparameter} & \textbf{Value} \\ \hline
Max episode length & $30\ min$  \\ \hline 
Num. action repeats & $4$ \\ \hline
Num. stacked frames & $1$ \\ \hline
Zero discount on life loss & $false$ \\ \hline
Random noops range & $30$ \\ \hline
Sticky actions & $false$ \\ \hline
Frames max pooled & 3 and 4\\ \hline
Grayscaled/RGB & Grayscaled \\ \hline
Action set & Full \\ \hline
\end{tabular}
\caption{Atari pre-processing hyperparameters.}
\label{table_pre-proc_dmlab}
\end{table}

\subsection{Control Suite pre-processing}
As mentioned in~\ref{sec:experiments}, we use no pre-processing on the frames received from the control environment.

\subsection{DmLab pre-processing}

\begin{table}[h]
\centering
\begin{tabular}{l|c}
\textbf{Hyperparameter} & \textbf{Value} \\ \hline
Num. action repeats & $4$ \\ \hline
Num. stacked frames & $1$ \\ \hline
Grayscaled/RGB & RGB \\ \hline
Image width & 96 \\ \hline
Image height & 72 \\ \hline
Action set & as in \cite{parisotto2020stabilizing} \\ \hline
\end{tabular}
\caption{DmLab pre-processing hyperparameters.}
\label{table_hyper_atari}
\end{table}

\subsection{Control environment discretization}
\label{app:control_discretization}
As mentioned, we discretize the space assigning two possibilities (1 and -1) to each dimension and taking the Cartesian product of all dimensions, which results in $2^n$ possible actions. For the {\ttfamily cartpole} tasks, we take a diagonal approach, utilizing each unit vector in the action space and then dividing each unit vector into 5 possibilities with the non-zero coordinate ranging from -1 to 1. The amount of actions this results in is outlined on Tab.~\ref{tab:control_actions}.

\begin{table}[h]
\centering
\begin{tabular}{l|c|c}
\textbf{Task} & \textbf{Action space size} & \textbf{Total amount of actions} \\ \hline
Acrobot & $1$ &  $2$ \\ \hline
Cartpole & $1$ & $5$ \\ \hline
Cup & $2$ & $4$ \\ \hline
Cheetah & $6$ & $64$ \\ \hline
Finger & $2$ & $4$ \\ \hline
Fish & $5$ & $32$ \\ \hline
Pendulum & $1$ & $2$ \\ \hline
Reacher & $2$ & $4$ \\ \hline
Swimmer & $5$ & $32$ \\ \hline
Walker & $6$ & $64$ \\ \hline
\end{tabular}
\caption{Control discretization action spaces.}
\label{tab:control_actions}
\end{table}

% \subsection{Hyperparameters Search Range}
% \label{app:search}
% The ranges we used to select the hyperparameters of \agent{} are displayed on Tab.~\ref{table_hyper_ranges}.
% \begin{table}[!ht]
% \centering
% \begin{tabular}{l|c}
% \textbf{Hyperparameter} & \textbf{Value} \\ \hline
% Q's $\lambda$ &  $\{0.8,\ 0.9\}$ \\ \hline
% Learning rate & $\{0.0001,\ 0.0003\}$ \\ \hline
% Contrastive loss weight & $\{0.01, 0.1, 1.0\}$ \\ \hline
% \end{tabular}
% \caption{HRange of hyperparameters usedsweeps for R2D2.}
% \label{table_hyper_ranges}
% \end{table}

\subsection{Hyperparameters Used}
\label{app:hypers_process}
We list all hyperparameters used here for completeness.

We started by optimizing the hyperparameters of GTrXL highlighted in bold in Tab \ref{tab:hyperparameters_R2D2} by doing a sweep over 10 Atari games: Seaquest, Qbert, Frostbite, Ms Pacman, Space Invaders, Gravitar, Solaris, Hero, Venture, Montezuma Revenge. Following this, the hyperparameters were kept fixed throughout all the experiments. 

Table \ref{tab:hyperparameters_R2D2} reports all the hyperparameters of the R2D2 experiments, both the fixed ones and the ones with the ranges over which we did the sweep. The fixed hyper-parameters were taken from \cite{kapturowski2018recurrent}. We then choose a set of hyper-paramters (both for the architecture and the algorithm) to sweep over to maximise the performance of gTrXL in this off-policy setting, given that there was not prior literature on this. The hyper-paramters over which we did the sweep for the algorithm were chose in accordance to \cite{kapturowski2018recurrent} and the related sweep. And for the architecture hyper-parameters we based our choice on \cite{parisotto2020stabilizing}. We believe that in this way we ensure to have a properly tuned baseline that enforces a fair comparison. Table \ref{tab:hyperparameters_R2D2_chosen} reports the chosen hyperparameters that we found to optimize the performance of GTrXL on the 10 Atari games. We then moved to CoBERL. CoBERL, in comparison to the baseline GTrXL has two extra hyperparamters: ‘Contrastive loss weight’ and ‘Contrastive loss mask rate’. The former was tuned, whereas the latter we kept equal to 0.15 as done in [11].
Consequently, we re-ran the same procedure as before, but we fixed all the previous hyperparameters optimized for GTrXL (see Tab.~\ref{tab:hyperparameters_R2D2_chosen} and we perform a grid search over the same 10 Atari games to find the value of ‘Contrastive loss weight’ that maximized performance. The values over which we did the search are 0.01, 0.1 and 1. We ended-up picking 1, although the difference between 0.1 and 1 was minimal. Table \ref{tab:table_hyper_ranges_rd2d2} reports the 2 extra parameters used for \agent{}.

For DmLAB we optimized the hyperparameters of GTrXL on all the 30 games. Table \ref{tab:hyperparameters_VMPO} reports both the fixed ones and the ones with the ranges over which. The fixed hyper-parameters were taken directly from \cite{parisotto2020stabilizing} and we sweep over ‘’Epsilon Alpha”, “Target Update Period” and 'Memory size' to make sure we maximised performance of this baseline. Table \ref{tab:hyperparameters_VMPO_chosen} reports the chosen hyperparameters that we found to optimize the performance of GTrXL on DmLAB. We then moved to CoBERL. Again, by keeping fixed all the previous hyperparameters optimized for GTrXL we perform a grid search over all the 30 games to find the value of ‘Contrastive loss weight’ that maximized performance. The values over which we did the search were 0.1 and 1. We did not find any significant difference between the two values, so we left it equal to 1 such that the two losses would have the same effect. Table \ref{tab:table_hyper_ranges_vmpo} reports the 2 extra parameters used for CoBERL and the reduced memory size, in accordance with our hypothesis that the LSTM on top of Transformer would help reducing the size of the memory especially in last.
\clearpage

\begin{small}
\begin{longtable}[h]{l|c}
\centering
\textbf{Hyperparameter} & \textbf{Value} \\ \hline
Optimizer & Adam \\ \hline
Learning rate & $\mathbf{\{0.0001,\ 0.0003\}}$ \\ \hline
Q's $\lambda$ & $\mathbf{\{0.8,\ 0.9\}}$ \\ \hline
Adam epsilon & $10^{-7}$  \\ \hline
Adam beta1 & $0.9$  \\ \hline
Adam beta2 & $0.999$  \\ \hline
Adam clip norm & $40$  \\ \hline
Q-value transform (non-DMLab) & $h(x) = sign(x)(\sqrt{|x| + 1} - 1) + \epsilon x$\\ \hline
Q-value transform (DMLab) & $h(x) = x$\\ \hline
Discount factor & $0.997$ \\ \hline
Batch size & $32$ \\ \hline
Trace length (Atari) & $80$ \\ \hline
Trace length (non-Atari) & $120$ \\ \hline
Replay period (Atari) & $40$ \\ \hline
Replay period (non-Atari) & $60$ \\ \hline
Replay capacity & $80000$ sequences \\ \hline
Replay priority exponent & $0.9$ \\ \hline 
Importance sampling exponent & $0.6$ \\ \hline 
Minimum sequences to start replay & $5000$ \\ \hline
Target Q-network update period & $400$ \\ \hline
Evaluation $\epsilon$ & $0.01$ \\ \hline
Target $\epsilon$ & $0.01$ \\ \hline
Number of layers & $\mathbf{\{6, 8, 12\}}$ \\ \hline
Memory size & $\mathbf{\{64, 128\}}$ \\ \hline
Number of heads & $\mathbf{\{4, 8\}}$ \\ \hline
Attention size & $\mathbf{\{64, 128\}}$ \\ \hline
\caption{GTrXL Hyperparameters used in all the R2D2 experiments with range of sweep.}
\label{tab:hyperparameters_R2D2}
\end{longtable}
\end{small}

\begin{small}
\begin{longtable}[h]{l|c}
\centering
\textbf{Hyperparameter} & \textbf{Value} \\ \hline
Optimizer & Adam \\ \hline
Learning rate & $\mathbf{0.0003}$ \\ \hline
Q's $\lambda$ & $\mathbf{0.8}$ \\ \hline
Adam epsilon & $10^{-7}$  \\ \hline
Adam beta1 & $0.9$  \\ \hline
Adam beta2 & $0.999$  \\ \hline
Adam clip norm & $40$  \\ \hline
Q-value transform (non-DMLab) & $h(x) = sign(x)(\sqrt{|x| + 1} - 1) + \epsilon x$\\ \hline
Q-value transform (DMLab) & $h(x) = x$\\ \hline
Discount factor & $0.997$ \\ \hline
Batch size & $32$ \\ \hline
Trace length (Atari) & $80$ \\ \hline
Trace length (non-Atari) & $120$ \\ \hline
Replay period (Atari) & $40$ \\ \hline
Replay period (non-Atari) & $60$ \\ \hline
Replay capacity & $80000$ sequences \\ \hline
Replay priority exponent & $0.9$ \\ \hline 
Importance sampling exponent & $0.6$ \\ \hline 
Minimum sequences to start replay & $5000$ \\ \hline
Target Q-network update period & $400$ \\ \hline
Evaluation $\epsilon$ & $0.01$ \\ \hline
Target $\epsilon$ & $0.01$ \\ \hline
Number of layers & $\mathbf{8}$ \\ \hline
Memory size & $\mathbf{64}$ \\ \hline
Number of heads & $\mathbf{8}$ \\ \hline
Attention size & $\mathbf{64}$ \\ \hline
\caption{GTrXL Hyperparameters choosen for all the R2D2 experiments.}
\label{tab:hyperparameters_R2D2_chosen}
\end{longtable}
\end{small}

\begin{table}[!ht]
\centering
\begin{tabular}{l|c}
\textbf{Hyperparameter} & \textbf{Value} \\ \hline
Contrastive loss weight & $1.0$ \\ \hline
Contrastive loss mask rate & $0.15$\\ \hline
\end{tabular}
\caption{Extra hyperparameters for CoBERL for the R2D2 experiments}
\label{tab:table_hyper_ranges_rd2d2}
\end{table}

\clearpage

\begin{longtable}[h]{l|c}
\centering
\textbf{Hyperparameter} & \textbf{Value} \\ \hline
Batch Size & 120 \\ \hline
Unroll Length & 95 \\ \hline
Discount & 0.99 \\ \hline
Target Update Period & $\mathbf{\{10,\ 20,\ 50\}}$ \\ \hline
Action Repeat & 4 \\ \hline
Initial $\eta$ & 1.0 \\ \hline
Initial $\alpha$ & 5.0 \\ \hline
$\epsilon_\eta$ & 0.1 \\ \hline
$\epsilon_\alpha$ & $\mathbf{\{0.001,\ 0.002\}}$ \\ \hline
Popart Step Size & 0.001 \\ \hline
Memory size & $\mathbf{\{256, 512\}}$ \\ \hline
\caption{GTrXL Hyperparameters used in all the VMPO experiments with range of sweep.}
\label{tab:hyperparameters_VMPO}
\end{longtable}

\begin{longtable}[h]{l|c}
\centering
\textbf{Hyperparameter} & \textbf{Value} \\ \hline
Batch Size & 120 \\ \hline
Unroll Length & 95 \\ \hline
Discount & 0.99 \\ \hline
Target Update Period & 50 \\ \hline
Action Repeat & 4 \\ \hline
Initial $\eta$ & 1.0 \\ \hline
Initial $\alpha$ & 5.0 \\ \hline
$\epsilon_\eta$ & 0.1 \\ \hline
$\epsilon_\alpha$ & 0.002 \\ \hline
Popart Step Size & 0.001 \\ \hline
Memory size & $\mathbf{512}$ \\ \hline
\caption{GTrXL Hyperparameters choosen for all the VMPO experiments.}
\label{tab:hyperparameters_VMPO_chosen}
\end{longtable}

\begin{table}[!ht]
\centering
\begin{tabular}{l|c}
\textbf{Hyperparameter} & \textbf{Value} \\ \hline
Contrastive loss weight & $1.0$ \\ \hline
Contrastive loss mask rate & $0.15$\\ \hline
Memory size & $256$ \\ \hline
\end{tabular}
\caption{Extra hyperparameters for CoBERL for the VMPO experiments}
\label{tab:table_hyper_ranges_vmpo}
\end{table}

\clearpage
\section{Additional ablations}
\label{app:additional-results}

Table~\ref{tab:gate_ablations} shows the results of several gating mechanisms that we have investigated. As we can observe the GRU gate is a clear improvement especially on DMLab, only being harmful in median on the reduced ablation set of Atari games.

\begin{table}[ht!]
\centering
\resizebox{\textwidth}{!}{%
\begin{tabular}{c|c|c|p{2.7cm}|p{2.7cm}|p{2.7cm}}
    \multicolumn{2}{c|}{} & \agent{} & 'Sum' gate & 'Concat' gate & w/o Gate\\ \hline
    \multirow{2}{*}{DMLab} & Mean   & $\mathbf{115.47\%\pm4.21\%}$  & $108.92\%\pm4.56\%$ & $105.93\%\pm4.82\%$ & $86.42\%\pm7.25\%$  \\ \cline{2-6} 
    & Median & $\mathbf{110.86\%}$ & $108.12\%$ & $107.82\%$ & $95.21\%$ \\ \hline
    \multirow{2}{*}{Atari} & Mean   & $\mathbf{726.34\%\pm44.06\%}$ & $548.66\%\pm11.16\%$ & $653.20\%\pm59.13\%$ & $591.33\%\pm91.25\%$   \\ \cline{2-6} 
    & Median & $270.03\%$ & $\mathbf{437.85}\%$ & $325.96\%$ & $320.09\%$  \\ \hline
    \end{tabular}
    }
    \caption{Gate ablations. Human normalized scores on Atari-57 ablation tasks and DMLab tasks. For the mean we include standard error over seeds.}
    \label{tab:gate_ablations}
\end{table}

\begin{table}[ht]
% \resizebox{\textwidth}{!}{%
\centering
\tiny
\begin{tabular}{c|c|c|c|c||c|c|c}
\hline
 DM Suite & \agent{} & R2D2-GTrXL & R2D2&  D4PG-Pixels & CURL & Dreamer & Pixel SAC\\
\hline
 acrobot swingup & \bf{355.25 $\pm$ 3.82} & 265.90 $\pm$ 113.27 & 327.16 $\pm$ 5.35 & 81.7 $\pm$ 4.4 & - & - & - \\
 fish swim & \bf{597.66 $\pm$ 60.09} & 82.30 $\pm$ 265.07 & 345.63 $\pm$ 227.44 & 72.2 $\pm$ 3.0 & - & - & - \\
 fish upright & \bf{952.60 $\pm$ 5.16} & 844.13 $\pm$ 21.70 & 936.09 $\pm$ 11.58 & 405.7 $\pm$ 19.6 & - & - & - \\
 pendulum swingup & \bf{835.06 $\pm$ 9.38} & 775.72 $\pm$ 50.06 & 831.86 $\pm$ 61.54 & 680.9 $\pm$ 41.9 & - & - & - \\
 swimmer swimmer6 & \bf{419.26 $\pm$ 48.37} & 206.95 $\pm$ 58.37 & 329.61 $\pm$ 26.77 & 194.7 $\pm$ 15.9 & - & - & - \\
 finger spin & \bf{985.96 $\pm$ 1.69} & 983.74 $\pm$ 10.2 & 980.85 $\pm$ 0.67 & \bf{985.7 $\pm$ 0.6} & 926 $\pm$ 45 & 796 $\pm$ 183 & 179 $\pm$ 166 \\
 reacher easy & \bf{984.00 $\pm$ 2.76} & \bf{982.60 $\pm$ 2.24} & \bf{982.28 $\pm$ 9.30} & 967.4 $\pm$ 4.1 & 929 $\pm$ 44 & 793 $\pm$ 164 & 145 $\pm$ 30 \\
 cheetah run & \bf{523.36 $\pm$ 41.38} & 105.23 $\pm$ 144.82 & 365.45 $\pm$ 50.40 & \bf{523.8 $\pm$ 6.8} & 518 $\pm$ 28 & \bf{570 $\pm$ 253} & 197 $\pm$ 15 \\
 walker walk & 781.16 $\pm$ 26.16 & 584.50 $\pm$ 70.28 & 687.18 $\pm$ 18.15 & \bf{968.3 $\pm$ 1.8} & 902 $\pm$ 43 & 897 $\pm$ 49 & 42 $\pm$ 12 \\
 ball in cup catch & 979.10 $\pm$ 6.12 & 975.26 $\pm$ 1.63 & \bf{980.54 $\pm$ 1.94} & 980.5 $\pm$ 0.5 & 959 $\pm$ 27 & 879 $\pm$ 87 & 312 $\pm$ 63 \\
 cartpole swingup & 812.01 $\pm$ 7.61 & 836.01 $\pm$ 4.37 & 816.23 $\pm$ 2.93 & \bf{862.0 $\pm$ 1.1} & 841 $\pm$ 45 & 762 $\pm$ 27 & 419 $\pm$ 40 \\
 cartpole swingup sparse & 714.80 $\pm$ 16.18 & 743.71 $\pm$ 9.27 & \bf{762.57 $\pm$ 6.71}  & 482.0 $\pm$ 56.6 & - & - & -\\
\hline
\end{tabular}
\caption{\small{Results on tasks in the DeepMind Control Suite. CoBERL, R2D2-GTrXL, R2D2, and D4PG-Pixels are trained on 100M frames, while CURL, Dreamer, and Pixel SAC are trained on 500k frames. We show these three other approaches as reference and not as a directly comparable baseline.}}
\label{tab:control_tab_baselines_full}
% }
\end{table}
\clearpage
\section{Game scores}
\label{app:scores}

\begin{table}[!ht]
\resizebox{\textwidth}{!}{%
\begin{tabular}{|c|c|c|c|c|}
\hline
Atari (ablation games) & \agent{} & R2D2-gTrXL & R2D2 & \agent{}-auxiliary loss \\
\hline
 beam rider & 29601.17 $\pm$ 11973.77 & \bf{65709.85 $\pm$ 29911.83} & 33938.58 $\pm$ 11340.33 & 43029.32 $\pm$ 47122.88 \\
 enduro & 2323.66 $\pm$ 31.42 & 2247.76 $\pm$ 139.16 & \bf{2338.98 $\pm$ 13.23} & 2250.01 $\pm$ 69.20 \\
 breakout & 409.87 $\pm$ 11.33 & 379.23 $\pm$ 32.37 & 363.43 $\pm$ 98.32 & \bf{420.72 $\pm$ 9.86} \\
 pong & 21.00 $\pm$ 0.00 & 20.95 $\pm$ 0.09 & \bf{21.00 $\pm$ 0.00} & 21.00 $\pm$ 0.00 \\
 qbert & 34000.79 $\pm$ 6074.45 & 22444.14 $\pm$ 4270.57 & 25375.24 $\pm$ 7451.33 & \bf{47741.30 $\pm$ 5288.90} \\
 seaquest & 164587.95 $\pm$ 122633.80 & \bf{310745.77 $\pm$ 67781.42} & 83490.38 $\pm$ 121434.46 & 174867.68 $\pm$ 123876.30 \\
 space invaders & \bf{37497.11 $\pm$ 9973.19} & 19113.28 $\pm$ 8512.98 & 4704.35 $\pm$ 2763.78 & 8783.28 $\pm$ 6450.09 \\
\hline
\end{tabular}

}
\end{table}

\begin{table}[!ht]
\resizebox{\textwidth}{!}{%
\begin{tabular}{|c|c|c|c|}
\hline
Atari (ablation games) & \agent{} & \agent{} with LSTM before & \agent{} w/o LSTM \\
\hline
 beam rider & 29601.17 $\pm$ 11973.77 & 35188.16 $\pm$ 23357.60 & \bf{62234.03 $\pm$ 12764.95} \\
 enduro & \bf{2323.66 $\pm$ 31.42} & 2321.02 $\pm$ 43.05 & 2286.74 $\pm$ 112.37 \\
 breakout & 409.87 $\pm$ 11.33 & \bf{418.13 $\pm$ 6.77} & 411.68 $\pm$ 13.62 \\
 pong & \bf{21.00 $\pm$ 0.00} & 21.00 $\pm$ 0.00 & 21.00 $\pm$ 0.00 \\
 qbert & \bf{34000.79 $\pm$ 6074.45} & 30055.59 $\pm$ 9892.69 & 31198.54 $\pm$ 5557.65 \\
 seaquest & 164587.95 $\pm$ 122633.80 & 172136.83 $\pm$ 119874.67 & \bf{192161.44 $\pm$ 94730.21} \\
 space invaders & \bf{37497.11 $\pm$ 9973.19} & 23746.32 $\pm$ 18961.05 & 10169.22 $\pm$ 12746.13 \\
\hline
\end{tabular}
}
\end{table}

\begin{table}[ht]
\resizebox{\textwidth}{!}{%
\begin{tabular}{|c|c|c|c|c|}
\hline
Atari (ablation games) & \agent{} & No Skip Conn. & Sum gate & Concat \\
\hline
 beam rider & 29601.17 $\pm$ 11973.77 & 52498.32 $\pm$ 2934.21 & \bf{72882.42 $\pm$ 21239.63} & 51371.38 $\pm$ 12560.94 \\
 enduro & \bf{2323.66 $\pm$ 31.42} & 2168.37 $\pm$ 362.99 & 2247.23 $\pm$ 82.22 & 2288.13 $\pm$ 58.59 \\
 breakout & \bf{409.87 $\pm$ 11.33} & 308.68 $\pm$ 141.28 & 403.36 $\pm$ 53.69 & 357.52 $\pm$ 72.63 \\
 pong & 21.00 $\pm$ 0.00 & 20.88 $\pm$ 0.18 & \bf{21.00 $\pm$ 0.00} & 21.00 $\pm$ 0.00 \\
 qbert & 34000.79 $\pm$ 6074.45 & 27660.71 $\pm$ 6830.06 & 33807.00 $\pm$ 4294.93 & \bf{43487.35 $\pm$ 14518.46} \\
 seaquest & 164587.95 $\pm$ 122633.80 & 118960.81 $\pm$ 112462.18 & 294976.55 $\pm$ 41827.55 & \bf{399817.75 $\pm$ 36729.78} \\
 space invaders & \bf{37497.11 $\pm$ 9973.19} & 22347.29 $\pm$ 15281.67 & 10387.59 $\pm$ 2858.63 & 20933.98 $\pm$ 13072.95 \\
\hline
\end{tabular}
}
\end{table}

\begin{table}[ht]
\resizebox{\textwidth}{!}{%
\begin{tabular}{|c|c|c|}
\hline
Atari (ablation games) & \agent{} with CURL & \agent{} with SimCLR \\
\hline
 beam rider & 25605.45 $\pm$ 12772.65 & \bf{36364.92 $\pm$ 27431.91} \\
 enduro & 2280.38 $\pm$ 152.75 & \bf{2343.53 $\pm$ 5.10} \\
 breakout & 337.39 $\pm$ 53.97 & \bf{397.40 $\pm$ 40.07} \\
 pong & 19.99 $\pm$ 2.02 & \bf{20.94 $\pm$ 0.11} \\
 qbert & 15935.73 $\pm$ 580.68 & \bf{24491.28 $\pm$ 14491.24} \\
 seaquest & 163715.19 $\pm$ 80515.52 & \bf{273070.09 $\pm$ 124529.37} \\
 space invaders & 14038.27 $\pm$ 16158.41 & \bf{22120.53 $\pm$ 17354.48} \\
\hline
\end{tabular}
}
\end{table}

% \begin{table}[ht]
% % \resizebox{\textwidth}{!}{%
% \centering
% \begin{tabular}{|c|c|c|c|}
% \hline
%  Control Suite & \agent{} & gTrXL & R2D2 \\
% \hline
%  acrobot swingup & \bf{355.25 $\pm$ 3.82} & 215.39 $\pm$ 122.82 & 327.16 $\pm$ 5.35 \\
%  fish swim & \bf{597.66 $\pm$ 60.09} &91.32 $\pm$ 277.15 & 345.63 $\pm$ 227.44 \\
%  fish upright & \bf{952.60 $\pm$ 5.16} & 849.52 $\pm$ 23.01 & 936.09 $\pm$ 11.58 \\
%  pendulum swingup & \bf{835.06 $\pm$ 9.38} &  743.65 $\pm$ 52.44 & 831.86 $\pm$ 61.54 \\
%  swimmer swimmer6 & \bf{419.26 $\pm$ 48.37} & 225.97 $\pm$ 60.67 & 329.61 $\pm$ 26.77 \\
%  finger spin & \bf{985.96 $\pm$ 1.69} & 977.41 $\pm$ 8.91 & 980.85 $\pm$ 0.67 \\
%  reacher easy & \bf{984.00 $\pm$ 2.76} & 981.64 $\pm$ 1.99 & 982.28 $\pm$ 9.30 \\
%  cheetah run & \bf{523.36 $\pm$ 41.38} & 115.15 $\pm$ 133.95 & 365.45 $\pm$ 50.40 \\
%  walker walk & 781.16 $\pm$ 26.16 & 595.96 $\pm$ 77.59 & 687.18 $\pm$ 18.15 \\
%  ball in cup catch & 979.10 $\pm$ 6.12 & 975.21 $\pm$ 1.77 & \bf{980.54 $\pm$ 1.94} \\
%  cartpole swingup & 812.01 $\pm$ 7.61 & \bf{837.31 $\pm$ 4.15} & 816.23 $\pm$ 2.93 \\
%  cartpole swingup sparse & 714.80 $\pm$ 16.18 & 747.94 $\pm$ 8.61 & \bf{762.57 $\pm$ 6.71} \\
% \hline
% \end{tabular}
% % }
% \end{table}

\begin{table}[ht]
\begin{tabular}{|c|c|c|c|}
\hline
Atari-57 & \agent{} & R2D2-gTrXL & R2D2 \\
\hline
 alien & \bf{10728.64 $\pm$ 6771.87} & 9593.30 $\pm$ 3226.26 & 8960.97 $\pm$ 4233.52 \\
 amidar & \bf{3089.76 $\pm$ 1028.80} & 2888.93 $\pm$ 1324.27 & 1980.60 $\pm$ 181.03 \\
 assault & 10055.81 $\pm$ 7856.45 & 7569.86 $\pm$ 4705.86 & \bf{18601.87 $\pm$ 6948.52} \\
 asterix & 732004.08 $\pm$ 342695.21 & 535334.42 $\pm$ 400048.35 & \bf{777461.28 $\pm$ 251661.19} \\
 asteroids & \bf{138026.82 $\pm$ 63177.27} & 108693.26 $\pm$ 22302.14 & 56238.62 $\pm$ 44595.74 \\
 atlantis & \bf{1062735.42 $\pm$ 47026.77} & 1005098.28 $\pm$ 36831.12 & 994240.79 $\pm$ 42158.91 \\
 bank heist & 1126.69 $\pm$ 620.85 & \bf{1131.62 $\pm$ 272.21} & 1110.12 $\pm$ 442.89 \\
 battle zone & 92092.43 $\pm$ 39167.40 & 77051.40 $\pm$ 45255.27 & \bf{94903.82 $\pm$ 20522.95} \\
 beam rider & 29601.17 $\pm$ 11973.77 & \bf{65709.85 $\pm$ 29911.83} & 33938.58 $\pm$ 11340.33 \\
 berzerk & 1288.99 $\pm$ 615.45 & 594.60 $\pm$ 117.78 & \bf{1314.18 $\pm$ 444.67} \\
 bowling & \bf{160.94 $\pm$ 53.41} & 49.23 $\pm$ 47.13 & 86.34 $\pm$ 27.65 \\
 boxing & 69.40 $\pm$ 51.78 & \bf{100.00 $\pm$ 0.00} & 99.28 $\pm$ 0.79 \\
 breakout & \bf{409.87 $\pm$ 11.33} & 379.23 $\pm$ 32.37 & 363.43 $\pm$ 98.32 \\
 centipede & 62816.39 $\pm$ 13516.24 & \bf{98575.88 $\pm$ 43443.70} & 75733.83 $\pm$ 19166.94 \\
 chopper command & \bf{614750.54 $\pm$ 283486.50} & 53852.34 $\pm$ 53906.17 & 26147.63 $\pm$ 14180.38 \\
 crazy climber & \bf{136884.73 $\pm$ 50591.49} & 117777.36 $\pm$ 18382.23 & 134610.46 $\pm$ 35360.45 \\
 defender & 347347.03 $\pm$ 90741.03 & 304551.03 $\pm$ 75288.86 & \bf{383931.18 $\pm$ 151219.51} \\
 demon attack & \bf{135904.86 $\pm$ 12721.85} & 115010.85 $\pm$ 26125.20 & 96371.00 $\pm$ 51049.84 \\
 double dunk & \bf{24.00 $\pm$ 0.00} & 21.25 $\pm$ 2.76 & 23.77 $\pm$ 0.33 \\
 enduro & 2323.66 $\pm$ 31.42 & 2247.76 $\pm$ 139.16 & \bf{2338.98 $\pm$ 13.23} \\
 fishing derby & 45.74 $\pm$ 9.05 & 39.83 $\pm$ 4.85 & \bf{47.47 $\pm$ 8.12} \\
 freeway & \bf{34.00 $\pm$ 0.00} & 34.00 $\pm$ 0.00 & 33.67 $\pm$ 0.42 \\
 frostbite & \bf{7877.53 $\pm$ 1919.45} & 6758.04 $\pm$ 2227.07 & 5850.66 $\pm$ 1443.76 \\
 gopher & 98169.07 $\pm$ 12126.03 & 84697.17 $\pm$ 26263.93 & \bf{106325.84 $\pm$ 17858.36} \\
 gravitar & \bf{6219.65 $\pm$ 484.85} & 5181.25 $\pm$ 1615.50 & 5186.02 $\pm$ 808.22 \\
 hero & \bf{20487.23 $\pm$ 288.81} & 16544.03 $\pm$ 3106.35 & 16764.65 $\pm$ 2575.03 \\
 ice hockey & \bf{24.64 $\pm$ 19.00} & 9.94 $\pm$ 12.69 & 21.38 $\pm$ 15.82 \\
 jamesbond & 5589.27 $\pm$ 4360.99 & \bf{7149.04 $\pm$ 3244.40} & 5872.92 $\pm$ 3432.57 \\
 kangaroo & 11703.01 $\pm$ 2892.01 & 11760.61 $\pm$ 3657.70 & \bf{13223.97 $\pm$ 2123.85} \\
 krull & 27738.24 $\pm$ 19282.30 & \bf{42886.34 $\pm$ 35329.57} & 19821.18 $\pm$ 17250.91 \\
 kung fu master & \bf{128429.78 $\pm$ 12649.00} & 64979.95 $\pm$ 25384.52 & 99741.61 $\pm$ 14185.70 \\
 montezuma revenge & \bf{502.27 $\pm$ 998.87} & 0.00 $\pm$ 0.00 & 240.00 $\pm$ 195.96 \\
 ms pacman & \bf{10567.67 $\pm$ 3692.12} & 9905.93 $\pm$ 884.03 & 9576.99 $\pm$ 2286.69 \\
 name this game & 26311.25 $\pm$ 6607.53 & \bf{26669.02 $\pm$ 3614.10} & 23889.30 $\pm$ 2608.30 \\
 phoenix & \bf{424708.26 $\pm$ 173336.21} & 283179.23 $\pm$ 94106.97 & 138558.03 $\pm$ 154881.71 \\
 pitfall & \bf{0.00 $\pm$ 0.00} & 0.00 $\pm$ 0.00 & 0.00 $\pm$ 0.00 \\
 pong & 21.00 $\pm$ 0.00 & 20.95 $\pm$ 0.09 & \bf{21.00 $\pm$ 0.00} \\
 private eye & 12034.62 $\pm$ 5963.85 & \bf{12690.27 $\pm$ 4372.38} & 3065.63 $\pm$ 5867.30 \\
 qbert & \bf{34000.79 $\pm$ 6074.45} & 22444.14 $\pm$ 4270.57 & 25375.24 $\pm$ 7451.33 \\
 riverraid & 22623.35 $\pm$ 3512.77 & 23692.95 $\pm$ 2100.11 & \bf{28147.78 $\pm$ 678.51} \\
 road runner & 212345.12 $\pm$ 192339.01 & \bf{542775.42 $\pm$ 58830.61} & 0.00 $\pm$ 0.00 \\
 robotank & \bf{80.69 $\pm$ 7.32} & 50.75 $\pm$ 31.02 & 63.82 $\pm$ 20.55 \\
 seaquest & 164587.95 $\pm$ 122633.80 & \bf{310745.77 $\pm$ 67781.42} & 83490.38 $\pm$ 121434.46 \\
 skiing & -29958.20 $\pm$ 4.99 & \bf{-21898.19 $\pm$ 9898.02} & -28243.31 $\pm$ 3453.55 \\
 solaris & \bf{6037.15 $\pm$ 3103.57} & 3675.08 $\pm$ 5575.93 & 2512.84 $\pm$ 1288.33 \\
 space invaders & \bf{37497.11 $\pm$ 9973.19} & 19113.28 $\pm$ 8512.98 & 4704.35 $\pm$ 2763.78 \\
 star gunner & \bf{103870.95 $\pm$ 23124.17} & 78559.08 $\pm$ 42110.81 & 88169.92 $\pm$ 13783.74 \\
 surround & 9.37 $\pm$ 0.45 & \bf{9.40 $\pm$ 0.51} & 8.60 $\pm$ 1.00 \\
 tennis & \bf{14.08 $\pm$ 11.58} & 9.20 $\pm$ 11.29 & -0.04 $\pm$ 0.08 \\
 time pilot & \bf{36085.82 $\pm$ 6934.00} & 14250.32 $\pm$ 1651.02 & 21051.93 $\pm$ 4156.28 \\
 tutankham & 33.93 $\pm$ 28.18 & 21.66 $\pm$ 13.11 & \bf{79.51 $\pm$ 40.10} \\
 up n down & 407873.64 $\pm$ 96615.12 & 237945.96 $\pm$ 101020.61 & \bf{438679.20 $\pm$ 38137.36} \\
 venture & \bf{1754.74 $\pm$ 229.65} & 1632.54 $\pm$ 247.91 & 1624.82 $\pm$ 100.65 \\
 video pinball & 248629.59 $\pm$ 231446.28 & 48305.17 $\pm$ 42432.44 & \bf{339391.16 $\pm$ 194812.76} \\
 wizard of wor & \bf{19454.60 $\pm$ 7061.60} & 14127.73 $\pm$ 9153.21 & 18270.16 $\pm$ 12166.04 \\
 yars revenge & \bf{385904.98 $\pm$ 264624.16} & 217128.25 $\pm$ 28248.52 & 248086.47 $\pm$ 162482.95 \\
 zaxxon & 20411.27 $\pm$ 12535.64 & 25063.44 $\pm$ 10385.30 & \bf{31679.84 $\pm$ 10447.57} \\
\hline

\end{tabular}
\end{table}

\begin{table}[ht]
\centering
\begin{tabular}{|c|c|c|}
\hline
 DmLab Levels & coberl & gtrxl \\ \hline
 rooms collect good objects test & $\mathbf{9.74\pm0.14}$ & $9.69\pm0.23$\\
 rooms exploit deferred effects test & $\mathbf{56.57\pm3.99}$ & $54.87\pm0.84$\\
 rooms select nonmatching object & $\mathbf{64.11\pm10.32}$ & $58.48\pm4.34$\\
 rooms watermaze & $\mathbf{56.61\pm1.49}$ & $51.72\pm3.41$ \\
 rooms keys doors puzzle & $33.99\pm5.44$ & $\mathbf{35.53\pm8.33}$ \\
 language select described object & $\mathbf{611.02\pm1.02}$ & $605.59\pm14.29$ \\
 language select located object & $\mathbf{606.05\pm3.73}$ & $584.23\pm3.22$ \\
 language execute random task & $\mathbf{232.67\pm28.07}$ & $198.45\pm15.96$ \\
 language answer quantitative question & $\mathbf{327.26\pm4.26}$ & $297.57\pm9.61$ \\
 lasertag one opponent large & $\mathbf{15.47\pm3.03}$ & $11.65\pm3.29$ \\
 lasertag three opponents large & $\mathbf{22.99\pm3.79}$ & $28.46\pm4.59$ \\
 lasertag one opponent small & $\mathbf{32.64\pm3.50}$ & $28.62\pm3.76$ \\
 lasertag three opponents small & $\mathbf{42.96\pm3.10}$ & $34.54\pm0.47$ \\
 natlab fixed large map & $\mathbf{46.03\pm8.85}$ & $43.46\pm14.38$ \\
 natlab varying map regrowth & $\mathbf{31.07\pm6.74}$ & $27.62\pm9.09$ \\
 natlab varying map randomized & $\mathbf{48.97\pm8.98}$ & $32.79\pm11.30$ \\
 platforms hard & $\mathbf{58.10\pm7.97}$ & $35.67\pm23.15$ \\
 platforms random & $\mathbf{85.15\pm1.54}$ & $69.01\pm1.04$ \\
 psychlab continuous recognition  & $\mathbf{58.91\pm2.17}$ & $54.96\pm3.87$ \\
 psychlab arbitrary visuomotor mapping & $51.56\pm0.44$ & $\mathbf{59.96\pm0.33}$ \\
 psychlab sequential comparison & $33.56\pm0.84$ & $\mathbf{35.89\pm0.64}$ \\
 psychlab visual search & $77.96\pm1.33$ & $76.92\pm0.82$\\
 explore object locations small & $\mathbf{83.07\pm5.65}$ & $71.85\pm6.23$ \\
 explore object locations large & $\mathbf{62.57\pm5.86}$ & $60.61\pm2.72$ \\
 explore obstructed goals small & $\mathbf{2.85\pm5.35}$ & $245.40\pm10.15$ \\
 explore obstructed goals large & $\mathbf{109.61\pm1.20}$ & $79.48\pm6.72$ \\
 explore goal locations small & $\mathbf{369.20\pm3.54}$ & $340.19\pm2.61$ \\
 explore goal locations large & $\mathbf{139.74\pm9.84}$ & $123.20\pm14.65$ \\
 explore object rewards few & $71.47\pm14.95$ & $69.02\pm0.95$ \\
 explore object rewards many & $\mathbf{104.62\pm2.63}$ & $94.97\pm1.8$ \\
\hline
\end{tabular}
\end{table}

\clearpage
\section{Learning Curves}
\label{app:learning-curves}

\subsection{Atari Learning curves}
\begin{figure}[h]
\label{fig:atari-curves}
\centering
  \includegraphics[width=15cm]{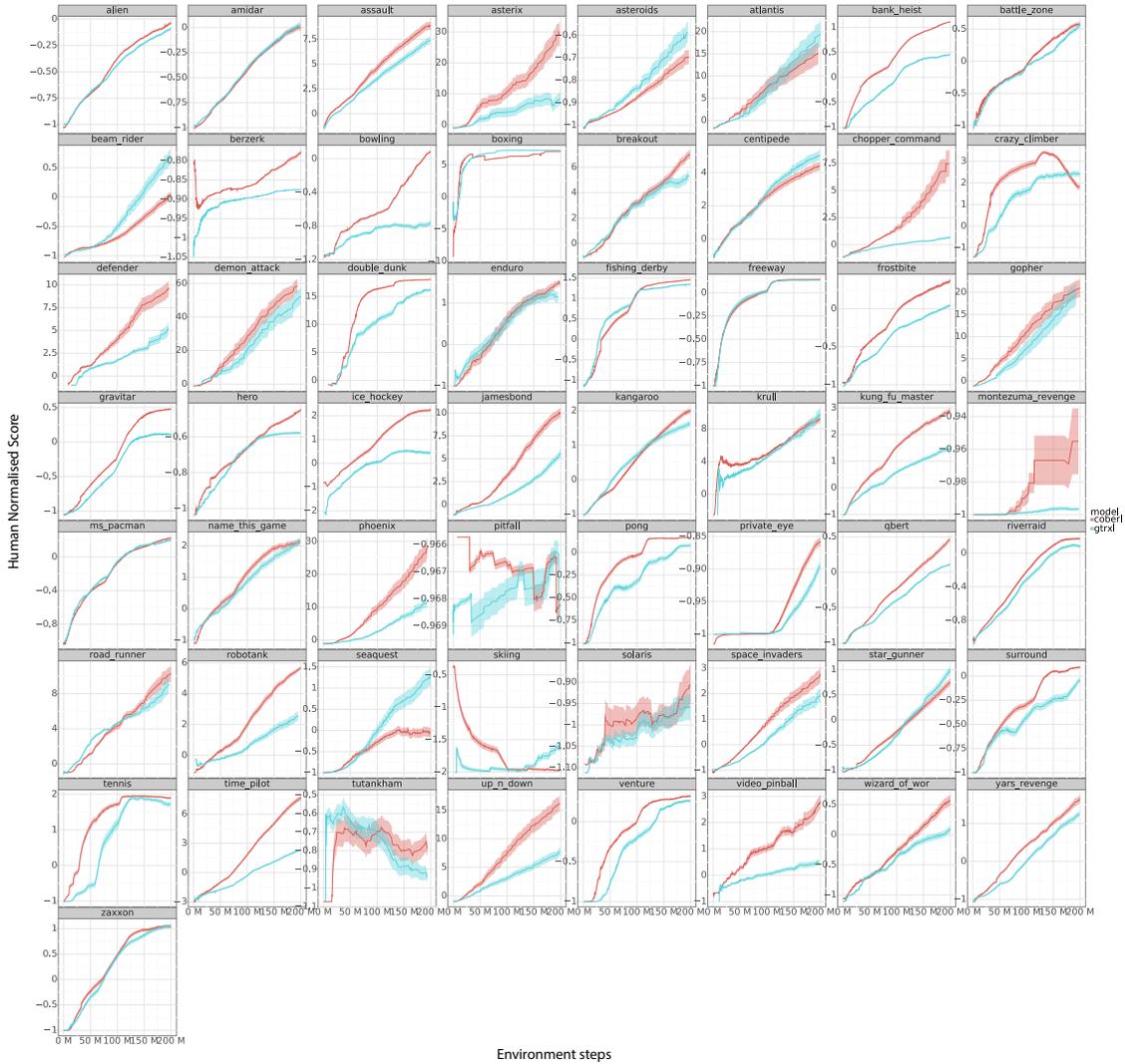}
  \vspace*{-0.7cm}
  \caption{\small{Learning Curves for Atari. The x-axis represents number of environment steps in millions. The y-axis represent the Human Normalised score. The error represents the 95\% confidence interval. Red is \agent{}, BLUE is GTrXL }}
  
\end{figure}

\clearpage

\subsection{DMControl Learning curves}
\begin{figure}[h]
\label{fig:dmcontrol-curves}
\centering
  \includegraphics[width=15cm]{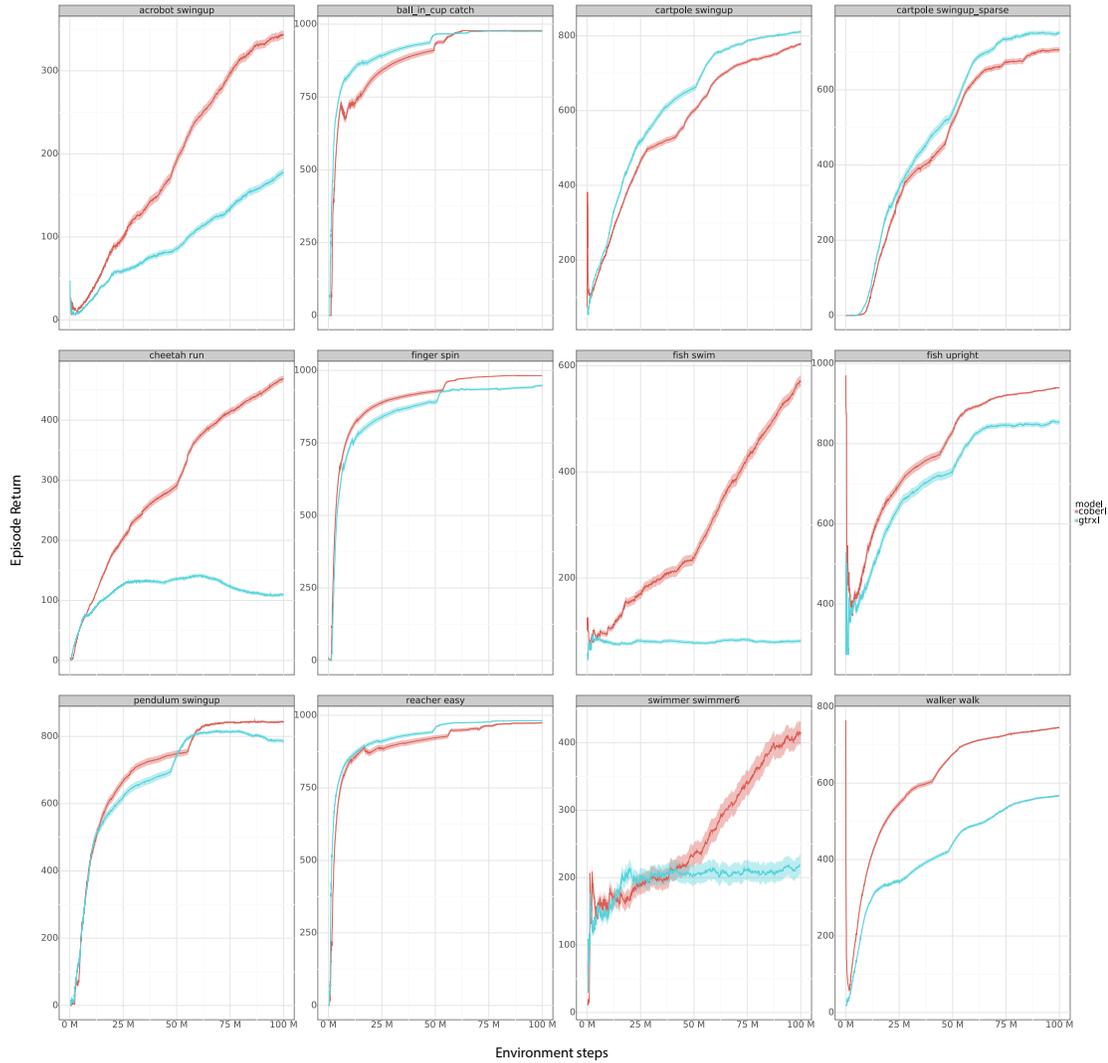}
  \vspace*{-0.7cm}
  \caption{\small{Learning Curves for DMControl. The x-axis represents number of environment steps in millions. The y-axis represent the Episode return. The error represents the 95\% confidence interval. Red is \agent{}, BLUE is GTrXL }}
  
\end{figure}

\clearpage

\subsection{DMLab Learning curves}
\begin{figure}[h]
\label{fig:dmlab-curves}
\centering
  \includegraphics[width=15cm]{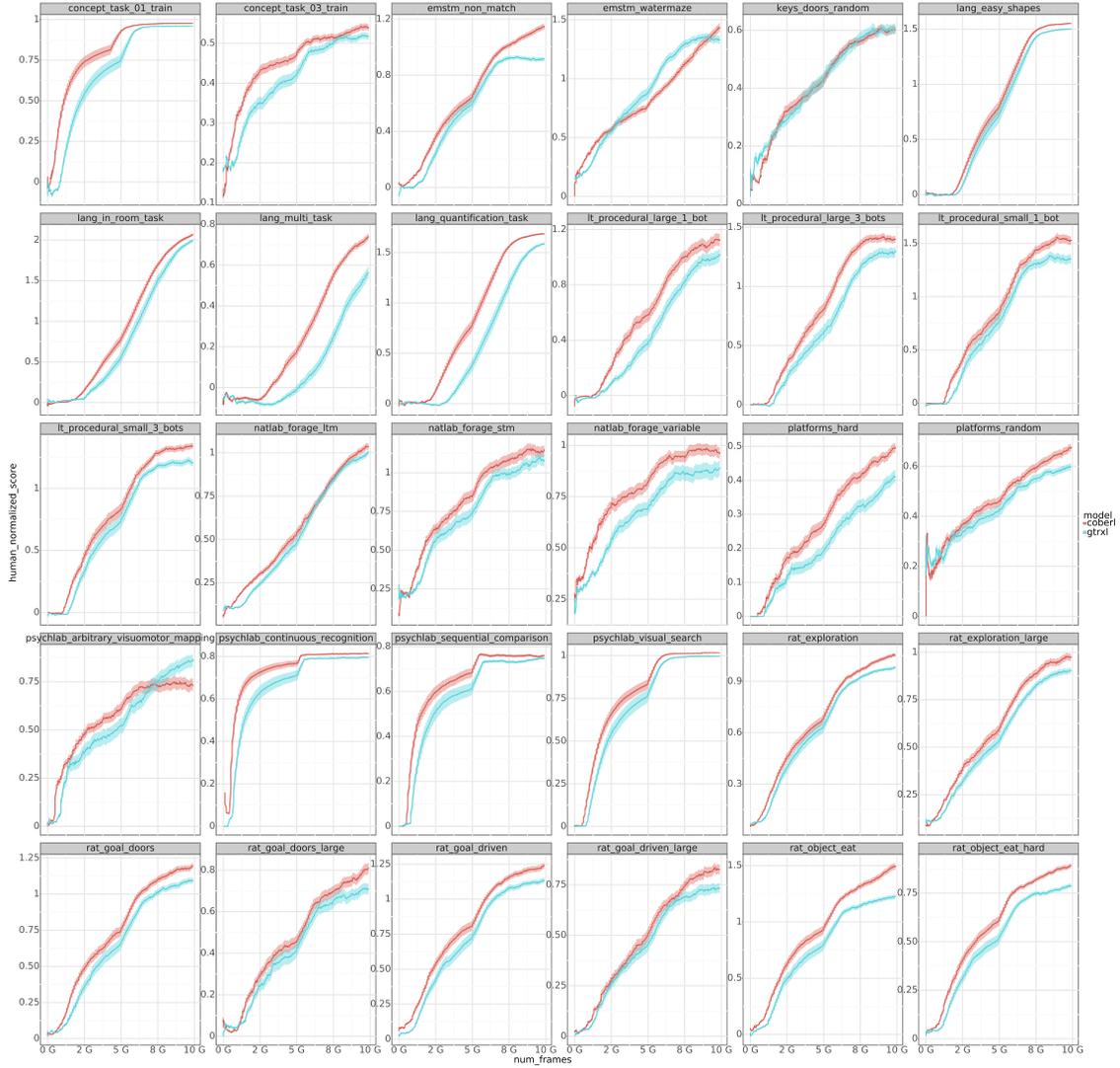}
  \vspace*{-0.7cm}
  \caption{\small{Learning Curves for DMLab. The x-axis represents number of environment steps in billions. The y-axis represent the Human Normalised score. The error represents the 95\% confidence interval. Red is \agent{}, BLUE is GTrXL }}
  
\end{figure}

\clearpage

\section{Licenses}
\label{app:licenses}

\textbf{The The Arcade Learning Environment}~\cite{bellemare2013arcade} is released as free, open-source software under the terms of the GNU General Public License. The latest version of the source code is publicly available at:
http://arcadelearningenvironment.org

\textbf{DeepMind Control Suite}~\cite{tassa2018deepmind} is released as free, open-source software under the terms of Apache-2.0 License. The latest version of the source code is publicly available at: \url{https://github.com/deepmind/dm_control/blob/master/dm_control/suite/README.md}

\textbf{DmLab}~\cite{beattie2016deepmind} is released as free, open-source software under the terms of Apache-2.0 License. The latest version of the source code is publicly available at: \url{https://github.com/deepmind/lab/tree/master/game_scripts/levels/contributed/dmlab30}
\section{Pseudo-code}
\label{app:pseudo-code}

\begin{algorithm}
\caption{Pseudo-code for CoBERL}\label{alg:coberl}
\begin{algorithmic}
\State  training\_iterations $\leftarrow$ 0
\State initialise\_weights(VisualEncoder, TransformerStack, Gate, ValueNetwork)

\While{ training\_iterations $ \leq $ max\_training\_iterations}
\State  sampled\_batch $\leftarrow$ sample\_experience()
\State encoded\_images $\leftarrow$ VisualEncoder(sampled\_batch)
\State encoded\_actions $\leftarrow$ OneHotActionEncoder(sampled\_batch)
\State combined\_inputs $\leftarrow$ concat(encoded\_images, previous\_rewards, encoded\_actions)
\State transformer\_inputs, transformer\_targets, mask $\leftarrow$ create\_masks\_targets(combined\_inputs)
\State output\_transformer\_contrastive $\leftarrow$ TransformerStack(transformer\_inputs, causal\_mask=False)
\State \textbf{contrastive\_loss} $\leftarrow$ compute\_aux\_loss(output\_transformer\_contrastive, transformer\_targets, mask)
\State output\_transformer\_RL $\leftarrow$ TransformerStack(transformer\_inputs, causal\_mask=True)
\State gated\_output $\leftarrow$ Gate(combined\_inputs, output\_transformer\_RL)
\State lstm\_output $\leftarrow$ LSTM(gated\_output)
\State combined\_inputs\_for\_value\_net  $\leftarrow$ concat(lstm\_output, combined\_inputs)
\State value\_estimation $\leftarrow$ ValueNetwork(combined\_inputs\_for\_value\_net)
\State \textbf{rl\_loss} $\leftarrow$ compute\_rl\_loss(value\_estimation, extra\_args)
\State \textbf{total\_loss} $\leftarrow$ rl\_loss + contrastive\_loss
\EndWhile
\end{algorithmic}
\end{algorithm}

\definecolor{codegreen}{rgb}{0,0.6,0}
\definecolor{codegray}{rgb}{0.5,0.5,0.5}
\definecolor{codepurple}{rgb}{0.58,0,0.82}
\definecolor{backcolour}{rgb}{0.95,0.95,0.92}

\lstdefinestyle{mystyle}{
    backgroundcolor=\color{backcolour},   
    commentstyle=\color{codegreen},
    keywordstyle=\color{magenta},
    numberstyle=\tiny\color{codegray},
    stringstyle=\color{codepurple},
    basicstyle=\ttfamily\footnotesize,
    breakatwhitespace=false,         
    breaklines=true,                 
    captionpos=b,                    
    keepspaces=true,                 
    numbers=left,                    
    numbersep=5pt,                  
    showspaces=false,                
    showstringspaces=false,
    showtabs=false,                  
    tabsize=2
}

\lstset{style=mystyle}
\clearpage

\textbf{Pseudo-code for the auxiliary loss calculation}
\begin{lstlisting}[language=Python]
def compute_aux_loss(input1, input2, mask_ext):
  
  batch_size, seq_dim, feat_dim = input1.shape
  input1 = reshape(input1, [-1, feat_dim])  
  input2 = reshape(input2, [-1, feat_dim])

  input1 = l2_normalize(input1, axis=-1)
  input2 = l2_normalize(input2, axis=-1)
  
  # Compute labels index    
  labels_idx = arange(input1.shape[0])
  labels_idx = labels_idx.astype(jnp.int32)
  # Compute pseudo-labels for contrastive loss.
  labels = one_hot(labels_idx, input1.shape[0] * 2)
  # Mask out the same image pair.
  mask = one_hot(labels_idx, input1.shape[0])
  # Compute logits.
  logits_11 = matmul(input1, jnp.transpose(input1))
  logits_22 = matmul(input2, jnp.transpose(input2))
  logits_12 = matmul(input1, jnp.transpose(input2))
  logits_21 = matmul(input2, jnp.transpose(input1))
  # Calculate invariance penalty.
  inv_penalty = kl_with_logits(  # [B * T]
      stop_gradient(logits_11), logits_22, axis=-1)
  inv_penalty += kl_with_logits(
      stop_gradient(logits_12), logits_22, axis=-1)
  inv_penalty += kl_with_logits(
      stop_gradient(logits_21), logits_11, axis=-1)
  inv_penalty += kl_with_logits(
      stop_gradient(logits_12), logits_21, axis=-1)
  inv_penalty = inv_penalty / 4.  # [B * T]

  logits_11 = logits_11 - mask * 1e9
  logits_22 = logits_22 - mask * 1e9

  logits_1211 = concatenate([logits_12, logits_11], axis=-1)
  logits_2122 = concatenate([logits_21, logits_22], axis=-1)

  loss_12 = -sum(labels * log_softmax(logits_1211), axis=-1)
  loss_21 = -sum(labels * log_softmax(logits_2122), axis=-1)

  loss = reshape(loss_12 + loss_21, [batch_size, seq_dim]) * mask_ext
  inv_penalty = reshape(inv_penalty, [batch_size, seq_dim]) * mask_ext

  loss = mean(loss + inv_penalty)

  return loss
\end{lstlisting}
\section{Area under the curve}
\label{app:auc}

For all the levels we calculated the AUC by integrating composite Simpson’s rule with a delta(x) of 5 steps. We use the \textit{intergate} package from scipy \citep{2020SciPy-NMeth}.
\section{Limitations and Future Work}
\label{app:limitations}
A limitation of our method is that it relies on single time step information to compute its auxiliary objective. Such objective could naturally be adapted to operate on temporally-extended patches, and/or action-conditioned inputs. Also, as done in the original BERT \citep{devlinetal2019bert}, it could be possible to add a CLS token at the beginning of each sequence sent to the Transformer and then train the CLS token with RL gradients. In this way it would be possible to directly use the embeddings of the CLS token as a sequence summary and hence provide more context to the policy estimation network.  We regard those ideas as promising future research avenues. 
\section{Extra information about the gate employed in equation \ref{eq:arch}}
\label{app:gate}

The gate we use in equation \ref{eq:arch} is derived directly from the one used in GTrXL \citep{parisotto2020stabilizing}. We report here the details for better clarity.

The gate is the defined in the following way:
\begin{equation}
    g(y, x) = (1-z) \odot y + z \odot \hat{h}
    \label{eq:gate-def}
\end{equation}

where:

\begin{equation}
    z=\sigma(W_z x + U_z y - b_g) 
    \label{eq:gate-z}
\end{equation}

\begin{equation}
    \hat{h}=\tanh(W_g x + U_g (r \odot y)) 
    \label{eq:gate-h}
\end{equation}

and:

\begin{equation}
    r=\sigma(W_r x + U_r y)
    \label{eq:gate-r}
\end{equation}

$W_z, U_z, W_g, U_g, W_r and U_r$ are set of linear weights and $b_g$ is a bias.

In our case x is the output of the transformer network and y is the output of the encoder network.

\clearpage
\section{Extra analysis}
\label{app:extra-analysis}

\begin{figure}[h]
\label{fig:extra-masking}
\centering
  \includegraphics[scale=0.5]{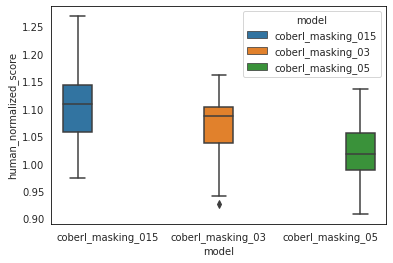}
  \caption{\small{Human normalised score as a function of the masking percentage on the input sequence. $coberl\_masking\_015$ relates to $15\%$ masking of the input sequence, $coberl\_masking\_03$ relates to ${30\%}$ masking of the input sequence and $coberl\_masking\_05$ relates to ${50\%}$ masking of the input sequence.}}
\end{figure}
We also attempted higher masking rate, but as shown in figure \ref{fig:extra-masking} they seem to perform worse, probably because high level of masking are reducing to much the number of frame present in the sequence, hence removing the information need to successfully perform the auxiliary task. 

\begin{figure}[h]
\label{fig:auc-100}
\centering
  \includegraphics[scale=0.5]{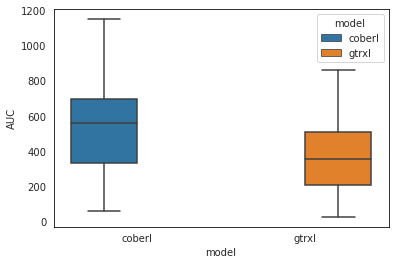}
  \caption{\small{AUC calculated at 100\% human score. }}
\end{figure}
Even when calulcate at 100\% human score, the AUC shows a significant advantage of \agent{} over gTrXL t(60)=2.616, p=0.0011. \agent{} M=545.18, SD=282.58; gTrXL M=374.49, SD=282.58.

\end{document}